\documentclass[11pt]{article}

\usepackage[preprint]{acl}

\usepackage{times}
\usepackage{latexsym}
 \usepackage{multirow}
\usepackage[T1]{fontenc}

\usepackage{xcolor}
\definecolor{posdelta}{HTML}{1A5490}   
\definecolor{negdelta}{HTML}{B33A1E}   

\usepackage[utf8]{inputenc}

\usepackage{microtype}

\usepackage{inconsolata}

\usepackage{booktabs}
\usepackage{graphicx}

\usepackage[table]{xcolor}
\usepackage[most]{tcolorbox}
\usepackage{inconsolata} 
\usepackage{listings}

\usepackage{listings}
\usepackage[most]{tcolorbox}
\usepackage{stfloats}

\newtcblisting{promptboxsingle}[1]{
  width=\linewidth,
  colback=gray!5,
  colframe=black!80,
  coltitle=white,
  colbacktitle=black!80,
  title=\textbf{#1},
  fonttitle=\normalsize\bfseries,
  boxrule=0.7pt,
  arc=1.5mm,
  left=1.5mm,
  right=1.5mm,
  top=1mm,
  bottom=1mm,
  listing only,
  listing options={
    basicstyle=\ttfamily\footnotesize,
    breaklines=true,
    breakatwhitespace=true,
    columns=fullflexible,
    keepspaces=true,
    showstringspaces=false,
    tabsize=2
  }
}

\newtcblisting{promptboxwide}[1]{
  width=0.95\textwidth,
  colback=gray!5,
  colframe=black!80,
  coltitle=white,
  colbacktitle=black!80,
  title=\textbf{#1},
  fonttitle=\large\bfseries,
  boxrule=0.8pt,
  arc=2mm,
  left=2mm,
  right=2mm,
  top=1mm,
  bottom=1mm,
  listing only,
    listing options={
      basicstyle=\ttfamily\footnotesize,
      breaklines=true,
      breakatwhitespace=true,
      columns=fullflexible,
      keepspaces=true,
      showstringspaces=false,
      tabsize=2
    }
}

\title{How Persona Conditioning Changes Open-Ended Task Execution: A Paired Evaluation of Behavioral Gains and Reliability Costs}

\author{
  \textbf{Yugyeong Ji}\textsuperscript{1} \quad
  \textbf{Changho Lee}\textsuperscript{2} \quad
  \textbf{Yerin Hwang}\textsuperscript{3} \\
  \textbf{Taegwan Kang}\textsuperscript{4} \quad
  \textbf{Jinsik Lee}\textsuperscript{4,*} \\[5pt]
  \textsuperscript{1}42dot \quad
  \textsuperscript{2}University of Michigan \\
  \textsuperscript{3}Dongguk University \quad
  \textsuperscript{4}LG AI Research \\[5pt]
  {\small
    \texttt{yugyeong.ji@42dot.ai} \quad
    \texttt{changhol@umich.edu}
  } \\
  {\small
    \texttt{yerinhwang@dongguk.edu}
  } \\
  {\small
    \texttt{\{teagwan.kang,jinsik.lee\}@lgresearch.ai}
  } \\
  {\footnotesize
    \textsuperscript{*}Corresponding author
  }
}

\begin{document}
\maketitle

\begin{abstract}
Large language models (LLMs) are often prompted with role- or persona-based instructions, yet persona conditioning has mainly been evaluated through character fidelity or aggregate benchmark accuracy. We study persona conditioning as a task-execution steering signal that changes how models interpret, structure, and sustain responses. Using paired instances derived from real user role-play logs and expanded into simulated multi-turn continuations, we compare \emph{with-persona} and \emph{no-persona} conditions while fixing task content, constraints, user simulation, and evaluation criteria. Across eight models, persona conditioning improves expert content depth and procedural fidelity, and these gains survive length control. At the same time, it lowers strict instruction-following accuracy by 17 to 30 points in every model evaluated, and increases unsupported real-world specificity through persona-induced elaboration. These results suggest that persona prompting is not a general-purpose performance improvement technique, but a selective task-execution policy shift with both behavioral gains and reliability risks.
\end{abstract}
\section{Introduction}
\label{sec:intro}

Large language models (LLMs) are increasingly prompted with role- or expertise-based instructions---such as responding as a doctor, senior engineer, or investor---under the expectation that role conditioning will elicit responses matching a target level of expertise and task framing.
Prior work has studied persona conditioning mainly from two perspectives: character role-playing, which evaluates fidelity to a target character~\citep{wang-etal-2024-rolellm, wang-etal-2024-incharacter, shao-etal-2023-character, zhou2025characterbench, lu-etal-2025-rolemrc}, and persona prompting, which examines whether expert-role instructions improve benchmark performance~\citep{xu2025expertpromptinginstructinglargelanguage, luz-de-araujo-etal-2025-principled, basil2025promptingsciencereport4, lutz2025promptmakespersonasystematic}.


\begin{figure}[t]
    \centering
    \includegraphics[width=\linewidth]{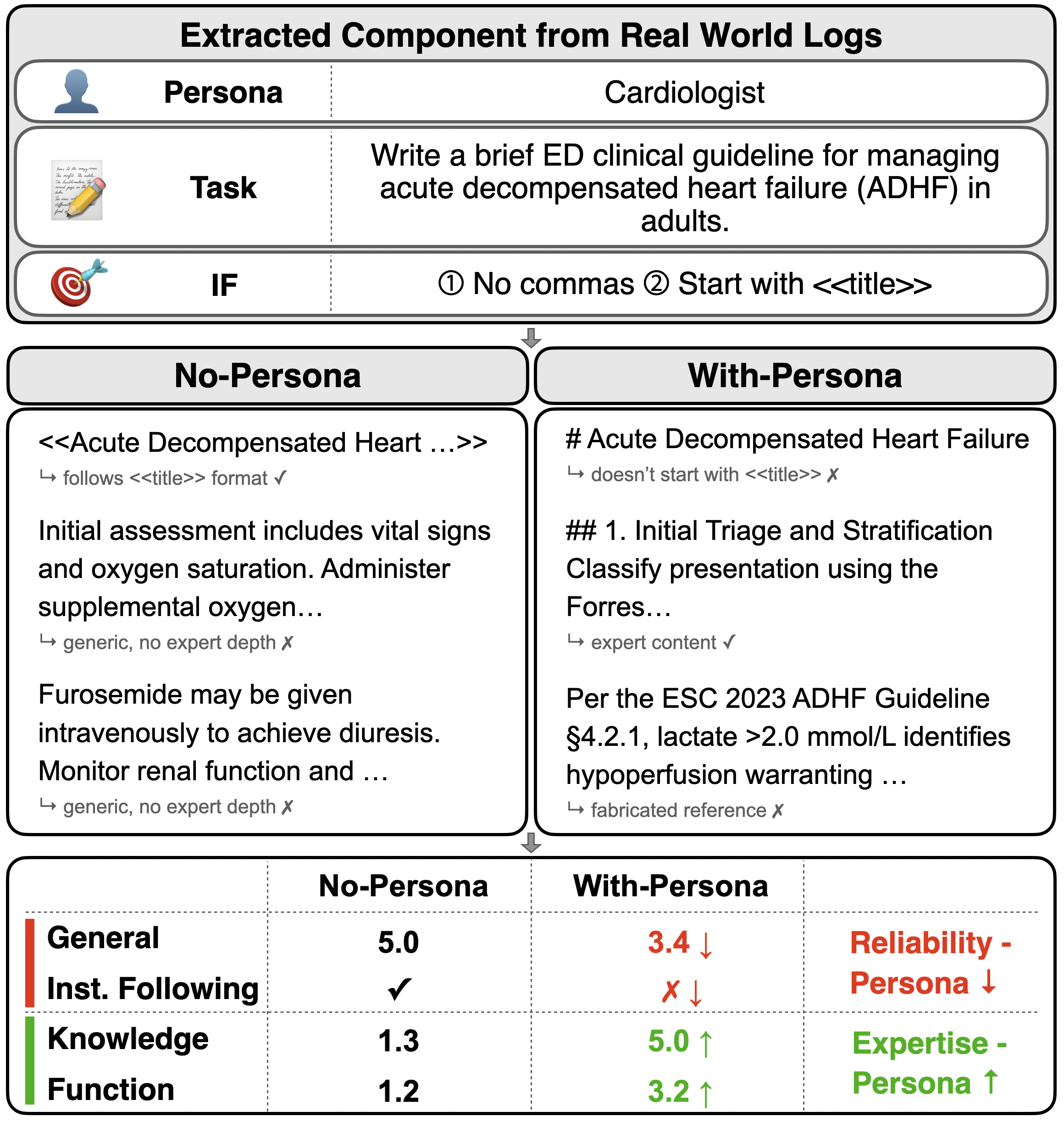} 
    \vspace{-5mm}
    \caption{
    Illustrative example of the persona-induced trade-off.
    Compared with the no-persona response, the with-persona response improves role-sensitive rubric scores while incurring costs in non-fabrication or instruction following.
    }
    \label{fig:main_figure}
    \vspace{-5mm}
\end{figure}

However, role-play in real user settings cannot be fully explained by either fidelity-centered or accuracy-centered perspectives. In real user conversation logs~\citep{zhao2024wildchat1mchatgptinteraction}, role-conditioned requests often use the specified role to shape how the task should be interpreted and what kind of answer should be produced. A request such as ``give feedback like a professor'' or ``analyze this as an experienced security engineer'' does not merely ask for a different tone; it specifies the criteria, priorities, and level of reasoning expected from the response. In this sense, personas function as task-framing signals that can shape how a model interprets the task, organizes its reasoning, and determines what counts as a useful answer.

To address this gap, we evaluate persona conditioning as a task-execution steering signal rather than only through character fidelity or aggregate benchmark performance. Specifically, we examine how role-based task framing translates into changes in task-execution behavior.
This steering can strengthen role-sensitive task execution by reshaping task interpretation, prioritization, and response structure, but it can also introduce reliability costs through unsupported specificity or conflicts with explicit constraints. Persona effects should therefore be understood as multidimensional behavioral shifts involving both role-sensitive gains and reliability costs~\citep{choi2024picleelicitingdiversebehaviors, miehling-etal-2025-evaluating}.

To operationalize this view, we design a multidimensional evaluation framework that separates role-sensitive behavioral gains from reliability costs. It evaluates responses along reliability, knowledge, function, and style dimensions, while separately measuring explicit instruction following with IFEval-compatible constraints. 
We instantiate this framework in a paired benchmark constructed from real user role-play interactions, where each task is evaluated both with and without an explicitly assigned persona while keeping task content, constraints, user simulation, and evaluation criteria fixed.

Our experiments show that persona conditioning induces a selective task-execution policy shift rather than a uniform performance improvement. As illustrated in Figure~\ref{fig:main_figure}, persona-conditioned responses improve role-sensitive dimensions such as expert reasoning, procedural organization, artifact usability, and role-appropriate register. At the same time, they reduce non-fabrication and strict instruction-following accuracy, often through unsupported specificity and violations of explicit constraints. These findings highlight persona prompting as a benefit--cost trade-off: it can strengthen task execution in role-sensitive ways while increasing factuality and constraint-following risks.

The contributions are summarized as follows:

\begin{itemize}
\setlength{\itemsep}{-2pt}
\setlength{\parsep}{0pt}
\setlength{\topsep}{2pt}
\item We frame persona conditioning as task-execution steering and design a multidimensional evaluation framework that jointly measures role-sensitive behavioral gains and baseline reliability costs.

\item We construct a paired benchmark from real user role-play interactions for controlled comparison of with-persona and no-persona conditions.

\item We show that persona conditioning induces a selective task-execution policy shift: it improves expert-oriented behavior and artifact usability, while reducing non-fabrication and strict instruction-following accuracy.
\end{itemize}
\section{Related Work}



\subsection{Character Role-Playing and Persona Fidelity}
A substantial body of work evaluates whether language models can faithfully adopt an assigned character or persona. These studies measure properties such as character fidelity, personality consistency, world knowledge consistency, stylistic consistency, and sustained role behavior across dialogue. Benchmarks such as RoleLLM~\citep{wang-etal-2024-rolellm}, InCharacter~\citep{wang-etal-2024-incharacter}, CharacterEval~\citep{tu-etal-2024-charactereval}, CharacterBench~\citep{zhou2025characterbench}, PersonaGym~\citep{samuel-etal-2025-personagym}, and RAIDEN~\citep{wu-etal-2025-raiden} primarily ask whether a model behaves consistently with a target identity or character description.

Our work differs in evaluation target. We do not primarily assess whether a response resembles a specific fictional character or faithfully reproduces a persona profile. Instead, we study how an assigned professional or social role changes the execution of a substantive downstream task. Persona embodiment remains one evaluated property, but we explicitly separate it from task-oriented dimensions that may exhibit role-sensitive gains, such as expert reasoning and procedural completion, as well as from condition-invariant reliability requirements that may reveal corresponding costs, such as unsupported real-world specificity and explicit instruction following~\citep{
zhou2023instructionfollowingevaluationlargelanguage,
harada-etal-2025-instructions,
deshpande-etal-2025-multichallenge,
jiang2024followbenchmultilevelfinegrainedconstraints}.

\subsection{Expert Persona Prompting}

A second line of research examines whether expert-role prompts improve task performance by eliciting more knowledgeable or structured responses. However, prior evaluations show that persona effects on benchmark accuracy can be small, heterogeneous across instances, and difficult to predict~\citep{basil2025promptingsciencereport4, xu2025expertpromptinginstructinglargelanguage, long-etal-2024-multi-expert, lutz2025promptmakespersonasystematic}. \citet{zheng-etal-2024-helpful} find small and difficult-to-predict effects of predefined persona labels on single-turn multiple-choice accuracy, while \citet{kim-etal-2025-persona} exploit instance-level variation through persona-conditioned versus neutral response selection to improve reasoning accuracy. Moving beyond accuracy, \citet{luz-de-araujo-etal-2025-principled} evaluate Expertise Advantage, Robustness, and Fidelity, highlighting that persona conditioning can affect multiple behavioral dimensions. More recently, PRISM~\citep{hu2026expertpersonasimprovellm} shows that expert persona effects depend on task type and model characteristics, yielding gains on alignment-dependent tasks but losses on knowledge-retrieval tasks. Collectively, these studies show that persona effects are heterogeneous: persona conditioning does not uniformly improve performance, and its effects cannot be adequately characterized by final-answer accuracy alone.

Our study builds on this literature but addresses a different setting and question. We focus on naturally occurring user-assigned personas attached to substantive open-ended tasks in real user--LLM interactions, rather than predefined personas applied to benchmark questions. Rather than asking whether persona prompting improves final-answer accuracy, we examine how persona conditioning selectively reshapes task execution, jointly measuring role-sensitive gains and reliability costs. We construct within-instance with-persona and no-persona comparisons of the same task and evaluate changes in expert content depth, procedural completion, unsupported real-world specificity, and strict instruction following. We further extend this comparison to simulated multi-turn trajectories.

\section{Evaluation Framework}
\label{sec:evaluation_framework}

\begin{table*}[t]
\centering
\small
\setlength{\tabcolsep}{4pt}
\renewcommand{\arraystretch}{1.15}
\begin{tabular}{p{0.17\linewidth} p{0.16\linewidth} p{0.24\linewidth} p{0.20\linewidth} p{0.10\linewidth}}
\toprule
Group & Category & Dimension & Evaluation focus & Eval method \\
\midrule
\multirow{4}{=}{\textbf{Baseline Reliability} \par \emph{condition-invariant requirements}}
& General (G)
& G1 Non-fabrication
& Grounds concrete real-world specifics in the prompt or context.
& Rubric \\
\cmidrule(l){2-5}
& \multicolumn{2}{l}{Instruction Following}
& Follows verifiable constraints.
& Rule-based \\
\midrule
\multirow{8}{=}{\textbf{Role-sensitive Behavior} \par \emph{shaped by task framing under a persona}}
& General (G)
& G2 Task consistency across turns
& Sustains task commitments.
& Rubric \\
\cmidrule(l){2-5}
& \multirow{2}{=}{Knowledge (A)}
& A1 Domain correctness
& Accurate domain claims.
& Rubric \\
& & A2 Expert content depth
& Expert-level reasoning.
& Rubric \\
\cmidrule(l){2-5}
& Function (B)
& B1 Procedural Fidelity
& Produces usable, procedurally complete artifacts.
& Rubric \\
\cmidrule(l){2-5}
& \multirow{3}{=}{Style (C)}
& C1 Register and genre framing
& Task-appropriate register.
& Rubric \\
& & C2 Persona embodiment fidelity
& Reflects assigned persona.
& Rubric \\
& & C3 Style consistency across turns
& Stable voice across turns.
& Rubric \\
\bottomrule
\end{tabular}
\caption{
Evaluation taxonomy. Baseline reliability dimensions are condition-invariant and include rule-based instruction following, whereas role-sensitive dimensions capture task-execution and style behaviors. Paired rubric dimensions (G1, G2, A1, A2, B1) use shared persona-neutral checklists, while Style (C1--C3) uses persona-anchored checklists and is evaluated only under persona conditioning.
}
\label{tab:behavioral_taxonomy}
\vspace{-5mm}
\end{table*}

\subsection{Overview}
\label{sec:evaluation_framework:overview}

Our framework evaluates role-playing as a change in task-execution behavior, rather than as character fidelity or style imitation alone. A persona can affect how the model interprets the task, prioritizes information, structures the response, and maintains behavior across turns, but these changes should not compromise baseline reliability.

We therefore separate persona effects into two types of dimensions, as summarized in Table~\ref{tab:behavioral_taxonomy}. \emph{Baseline reliability} dimensions capture condition-invariant requirements, including G1 non-fabrication and \textsc{IFEval}-compatible instruction following. \emph{Role-sensitive task execution and adherence} dimensions capture behaviors that may be reweighted by persona-induced task framing, including knowledge, function, style, and task consistency across turns. We define these dimensions in Sections~\ref{sec:evaluation_framework:baseline_reliability} and~\ref{sec:evaluation_framework:role_sensitive}, and describe the evaluation methodology in Section~\ref{sec:evaluation_framework:methodology}.

\subsection{Baseline Reliability Dimensions}
\label{sec:evaluation_framework:baseline_reliability}

Baseline reliability dimensions capture requirements that should remain stable regardless of whether a persona is provided. They are not intended to reward persona-specific behavior; instead, they measure whether the model preserves basic reliability while performing the task.

\paragraph{Non-fabrication (G1).}
G1 measures whether the response introduces concrete real-world specifics---citations, studies, URLs, statistics, named entities, and attributed quotations---that it asserts without support. The evaluation performs no retrieval or external verification, so G1 captures unsupported specificity rather than verified falsehood. Like the instruction-following constraints, G1 is condition-invariant: a persona-conditioned response and its no-persona counterpart are held to the same standard.

\paragraph{Instruction following.}
We also evaluate compliance with explicit, deterministically verifiable constraints using \textsc{IFEval}-compatible instructions. These constraints include JSON output, number of sections, word count, required or forbidden keywords, response language, and fixed ending phrases. Like G1, these constraints define condition-invariant requirements: the same constraints are applied to both the with-persona and no-persona conditions. This allows us to test whether persona conditioning interferes with explicit instruction compliance while changing task-execution behavior.

\subsection{Role-Sensitive Behavioral Dimensions}
\label{sec:evaluation_framework:role_sensitive}
Role-sensitive behavioral dimensions capture aspects of task execution that may change depending on how a persona frames the task. Unlike baseline reliability dimensions, these dimensions are not purely condition-invariant constraints. Instead, they measure how the model performs the task: what knowledge it prioritizes, how deeply it reasons, how concretely it constructs the output, whether it sustains task commitments across turns, and whether it adopts an appropriate role-conditioned style.

\paragraph{Task Consistency Across Turns (G2).}
G2 measures whether task-related commitments established in earlier turns are maintained in later turns. These commitments may include factual claims, decisions, content focus, level of detail, reasoning approach, and output format. Although G2 belongs to the General category, we treat it as part of role-sensitive task execution because maintaining task commitments over a trajectory is central to carrying out the task consistently. In this sense, G2 bridges reliability and task execution: it reflects a general consistency requirement, but its realization depends on the task framing established by the interaction.

\paragraph{Knowledge (A1, A2).}
Knowledge dimensions evaluate the content-level quality of the response. A1 Domain Correctness measures whether the response provides accurate domain claims, technical facts, rules, calculations, procedural steps, and named entities. A2 Expert Content Depth measures whether the response goes beyond generic explanation and reflects expert-level reasoning, such as task-appropriate prioritization, awareness of trade-offs and edge cases, and calibrated use of domain-specific concepts.

\paragraph{Function (B1).}
Function dimensions evaluate whether the response carries out the requested task in a procedurally complete and usable manner. B1 Procedural Fidelity measures whether outputs such as code, tables, checklists, plans, scripts, tutorials, or professional documents follow the task-appropriate procedure, include the necessary components, and produce a complete artifact rather than a skeletal outline or placeholder.

\paragraph{Style (C1, C2, C3).}
Style dimensions evaluate whether the model adopts and sustains the communicative behavior implied by the persona and task. C1 Register and Genre Framing measures whether the response uses the tone, register, genre, and terminology density appropriate for the task and audience. C2 Persona Embodiment Fidelity measures whether the response reflects the perspective, experience, value judgments, and persona-specific anchors of the assigned role, rather than relying on a generic role archetype. C3 Style Consistency Across Turns measures whether the voice, tone, register, and audience-calibrated vocabulary established in earlier turns are maintained throughout the dialogue.

\subsection{Evaluation Methodology}
\label{sec:evaluation_framework:methodology}

We use two evaluation procedures depending on the type of dimension. 
Instruction following is evaluated deterministically using the official \textsc{IFEval} logic, which checks whether each response satisfies the injected verifiable constraints, such as format, length, keyword, language, and fixed-ending requirements.


All remaining dimensions are evaluated using an LLM-judge-based checklist rubric. Rubric categories are selected from the persona-neutral first-turn user query. For paired dimensions (G1, G2, A1, A2, B1), task-specific checklist items are generated from this persona-neutral query and applied identically to both conditions.

Style dimensions (C1--C3) are handled separately: their checklist items are generated from the with-persona query and evaluated only under the with-persona condition. Accordingly, Style scores are reported as absolute scores rather than paired persona effects. Each checklist item is evaluated as pass or fail.

Most rubric-based dimensions are evaluated at the response level. 
However, G2 Task Consistency Across Turns and C3 Style Consistency Across Turns are evaluated over the full multi-turn dialogue, since they measure whether task commitments and style are sustained across the interaction. 
For each rubric-based sub-dimension, the checklist pass rate is converted into a 1--5 score.
\section{Paired Benchmark Construction}
\label{sec:benchmark_construction}

\begin{table*}[!t]
\centering
\small
\setlength{\tabcolsep}{4.5pt}
\renewcommand{\arraystretch}{1.10}

\begin{tabular}{@{}l rr rrrr@{}}
\toprule
&
\multicolumn{2}{c}{Baseline Reliability}
&
\multicolumn{4}{c}{Role-Sensitive Task Execution} \\
\cmidrule(lr){2-3}
\cmidrule(lr){4-7}

Model
& \shortstack{G1 Non-\\fabrication}
& \shortstack{Instruction\\Following}
& \shortstack{A1 Domain\\Correctness}
& \shortstack{A2 Expert\\Content Depth}
& \shortstack{B1 Procedural\\Fidelity}
& \shortstack{G2 Task Consistency\\Across Turns} \\
\midrule

Qwen3.5-4B
& -2.45 & -24.53 & +6.13 & +19.81 & +13.74 & +6.95 \\

Qwen3.5-9B
& -1.60 & -26.95 & +6.63 & +18.45 & +12.50 & +6.13 \\

Qwen3.5-35B-A3B
& +0.05 & -27.06 & +8.04 & +19.61 & +12.51 & +5.94 \\

Qwen3.5-397B-A17B
& -0.67 & -28.93 & +9.07 & +19.27 & +16.08 & +7.59 \\

GPT-5.4
& +0.91 & -20.06 & +4.53 & +7.70 & +10.24 & +4.34 \\

Claude-Sonnet-4.6
& -0.95 & -16.97 & -4.61 & -0.22 & -4.70 & +0.09 \\

Gemini-3-Flash
& -1.59 & -29.29 & +1.58 & +10.54 & +3.85 & +2.27 \\

Llama-3.3-70B
& -0.84 & -30.14 & +0.46 & +1.02 & +3.83 & -4.80 \\

\bottomrule
\end{tabular}

\vspace{-1mm}
\caption{
Persona-induced changes across all evaluated models.
Values report $\Delta_\%$, the with-persona minus no-persona change
expressed in percentage points relative to each metric scale.
Positive values indicate higher scores under persona conditioning,
whereas negative values indicate lower scores.
Full condition scores and additional paired statistics are reported in
Appendix~\ref{app:detailed_paired_results}; pairwise A/B results are reported
separately in Appendix~\ref{app:ab_evaluation}.
}
\label{tab:main_result_v2}
\vspace{-4mm}
\end{table*}


We construct a paired multi-turn evaluation set from real user--LLM conversations in ShareGPT90K\footnote{\url{https://huggingface.co/datasets/liyucheng/ShareGPT90K}} and WildChat\footnote{\url{https://huggingface.co/datasets/allenai/WildChat}}, where each pair differs only in whether role conditioning is applied to the same task.

Benchmark construction consists of five stages. First, we identify conversations in which the user explicitly assigns a role or expertise and asks a downstream task. Second, we extract the user-specified role as the \texttt{persona} and the following substantive request as the \texttt{task}. Third, we rewrite each instance into \texttt{with\_persona} and \texttt{no\_persona} conditions while preserving the same task content, language, length, and output constraints. We remove instances with residual role cues or role-dependent tasks to avoid baseline contamination. Fourth, we inject the same deterministically verifiable constraints into both conditions for symmetric instruction-following evaluation. Finally, we expand each pair into a 3--5 turn conversation between the evaluated assistant model and a fixed user simulator (Qwen3.5-397B-A17B) for trajectory-level evaluation. Both conditions use the same simulator model and generation procedure, but the simulator observes condition-specific conversation histories when generating follow-up turns. Thus, while the first-turn prompts differ only in role assignment, later-turn comparisons characterize the resulting interaction trajectories rather than fixed-history causal effects.


This construction keeps task content, output constraints, the user simulator, and evaluation criteria fixed across conditions, leaving the role assignment as the only experimental variable. Appendix~\ref{app:benchmark_construction} provides additional construction details and descriptive statistics of generated conversations.
\section{Experiments}
\label{sec:experiments}
\subsection{Metrics}
\label{sec:experiments:metrics}

We report paired metrics comparing the with-persona and no-persona conditions. For each rubric-based sub-dimension, we compute the mean 1--5 score for each condition and the paired difference $\Delta = s_{\text{with-persona}} - s_{\text{no-persona}}$. For compact cross-model comparison, the main results report $\Delta_\%$, the corresponding percentage-point change relative to each metric scale. Full condition scores, raw differences, and headroom-normalized changes are reported in Appendix~\ref{app:detailed_paired_results}.
For instruction following, we report prompt-level strict accuracy under the deterministic \textsc{IFEval} evaluation. As with the rubric-based dimensions, paired differences, effect sizes, and significance tests are computed over matched with-persona and no-persona instances.

We additionally define a task-execution aggregate by averaging A1 Domain Correctness, A2 Expert Content Depth, B1 Procedural Fidelity, and G2 Task Consistency Across Turns. Although G2 belongs to the General category, we include it in this aggregate because sustaining task-related commitments across turns is central to consistent task execution. Since Style (C) measures role adherence under an assigned persona, we report C1, C2, and C3 as absolute with-persona scores rather than paired gains.

Separately, we report pairwise A/B preference results as Persona Win\%, the proportion of comparisons in which the with-persona response is preferred over the corresponding no-persona response for each evaluation dimension. These results are reported in
Appendix~\ref{app:ab_evaluation}.

\subsection{Models}
\label{sec:experiments:models}

We evaluate four Qwen3.5 models as a within-family scale axis:
Qwen3.5-4B, 9B, 35B-A3B, and 397B-A17B~\citep{qwen2026qwen35} to analyze how scale affects behavioral gains and reliability costs within the same model family.

For cross-family comparison, we additionally evaluate Llama-3.3-70B~\citep{meta2024llama33}, Gemini 3 Flash Preview~\citep{google2026gemini3flash}, GPT-5.4~\citep{openai2026gpt54}, and Claude 4.6 Sonnet~\citep{anthropic2026claude46sonnet}. This allows us to examine whether the role-sensitive gains and reliability costs observed in the Qwen family also hold across other model families and closed frontier settings. 
Qwen3.5-397B-A17B also serves as the fixed user simulator and primary LLM judge. Because it is included as a subject model, we additionally assess robustness through cross-family judge swaps and analyses excluding this model (Section~\ref{sec:results} and Appendix~\ref{app:judge_swap}).
\section{Results}
\label{sec:results}

\paragraph{Overall results.}

As shown in Table~\ref{tab:main_result_v2}, persona conditioning does not yield uniform performance improvements across all evaluation dimensions.
Instead, it produces a benefit--cost trade-off in which gains in role-sensitive task execution are accompanied by reliability costs. This pattern is also reflected in persona win rates: persona-conditioned responses win more often on role-sensitive dimensions such as expert content depth and procedural fidelity, but less often on non-fabrication. An aggregate visualization of these behavioral gains and reliability costs across models is provided in Appendix~\ref{app:aggregate_visualization}. Persona-conditioned responses generally improve over the no-persona baseline in task-execution score, suggesting that personas can strengthen domain correctness, expert problem framing, procedural organization, and multi-turn task consistency. In contrast, non-fabrication and \textsc{IFEval} strict accuracy tend to decrease under the same condition, indicating increased unsupported real-world specificity and violations of explicit constraints.

Style scores under the with-persona condition show high C1 register and C3 
consistency but comparatively lower C2 persona embodiment (Appendix~\ref{app:extended_results}), 
suggesting that models reproduce a role-appropriate register more readily than 
the task-specific perspective of the assigned role. Given the low annotator 
agreement on C1 and C3 (Section~\ref{sec:analysis:human_validation}), we report 
these dimensions descriptively rather than as quantitative evidence.

\paragraph{Sub-dimension results.}

At the sub-dimension level, the positive effects of persona conditioning are concentrated in expert content depth, procedural organization, and style-related dimensions. In A2, persona-conditioned responses exhibit stronger expert framing and task-specific reasoning. In B1, they more often provide structured procedural organization and actionable response formats.

In contrast, G1 reveals an increase in unsupported claims and fabricated specificity, as persona-conditioned responses often become more concrete and assertive. This cost operates largely through the increased elaboration that persona conditioning induces rather than independently of response length (Appendix~\ref{app:length}). These findings indicate that persona conditioning affects dimensions asymmetrically, strengthening expert-oriented behavior while expanding the surface area for unsupported specifics.

\paragraph{Instruction-following results.}
In the instruction-following evaluation, persona-conditioned responses show lower \textsc{IFEval} strict accuracy than the no-persona baseline across all evaluated models. However, this instruction-following cost is not distributed uniformly across constraint types. Constraint-level analysis shows that the largest degradation occurs in content-level constraints, which directly regulate the content within the response body. By contrast, wrapping-level constraints, which govern the outer structure of the response, show comparatively smaller changes. In other words, persona conditioning does not uniformly collapse strict formatting behavior. Instead, it tends to selectively impair content-level constraints while relatively preserving wrapper-level constraints.

\paragraph{Robustness of the reported effects.}
Because most dimensions are scored by an LLM judge that shares a model family with one subject model, we verify that the reported effects are not artifacts of the evaluation setup. Re-scoring 200 paired instances with Gemini, GPT, and Claude judges preserves positive A2 and B1 effects and negative G1 effects, whereas A1 and G2 change sign across evaluator families; we therefore report A1 and G2 as specific to the primary judge. Excluding the self-judged subject model preserves the pattern (A2 $+0.549$, B1 $+0.371$, instruction following $-0.250$). Under length matching and regression adjustment, A2 and B1 remain significantly positive, while the G1 decrease does not survive adjustment; we accordingly interpret G1 as a cost mediated by persona-induced elaboration rather than a length-independent factuality penalty. The instruction-following decline is unaffected by length adjustment (odds ratio $0.124 \rightarrow 0.127$) and is evaluated deterministically. Appendices~\ref{app:judge_swap} and~\ref{app:length} report these analyses in full.
\section{Analysis}
\label{sec:analysis}
\begin{figure}[t]
    \centering
    \includegraphics[width=0.95\linewidth]{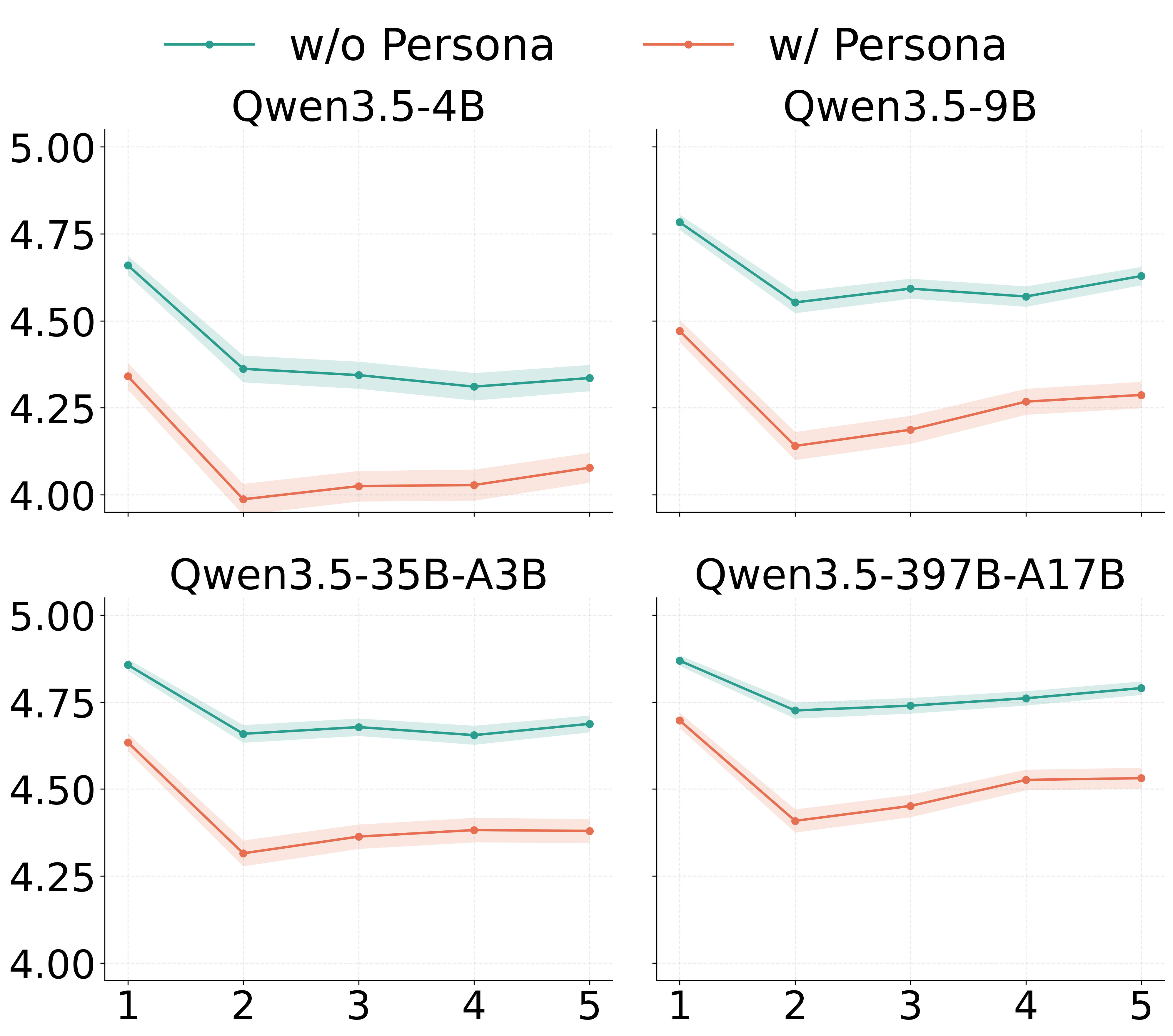}
    \vspace{-2mm}
    \caption{
    Turn-level G1 non-fabrication scores under cumulative dialogue context. Each panel plots the G1 score at turn $t$, computed using the dialogue history from turn 1 to turn $t$.
    }
    \label{fig:turn_accumulation}
    \vspace{-5mm}
\end{figure}
\begin{table}[t]
\centering
\small
\setlength{\tabcolsep}{4pt}
\begin{tabular}{lcccccc}
\toprule
& \multicolumn{3}{c}{Item $\kappa$} & \multicolumn{3}{c}{Score $\rho$} \\
\cmidrule(r){2-4}\cmidrule(l){5-7}
Dimension & H1 & H2 & H3 & H1 & H2 & H3 \\
\midrule
A2 Expert Depth    & .755 & .784 & .763 & .889 & .838 & .838 \\
B1 Procedural Fid. & .728 & .727 & .784 & .817 & .845 & .860 \\
G1 Non-fabrication & .586 & .491 & .632 & .588 & .545 & .617 \\
\midrule
Overall            & .693 & .596 & .700 & .771 & .703 & .757 \\
\bottomrule
\end{tabular}
\caption{Human validation of the scoring protocol on 100 paired instances, comprising 2{,}443 checklist-item judgments and 566 dimension-level scores.}
\label{tab:human_validation}
\end{table}
\begin{table*}[!t]
\centering
\small
\setlength{\tabcolsep}{5pt}
\renewcommand{\arraystretch}{1.12}
\begin{tabular}{p{0.48\textwidth}p{0.48\textwidth}}
\toprule
\textbf{w/o P} & \textbf{w/ P: ``data scientist''} \\
\midrule

\multicolumn{2}{l}{\textbf{User query}} \\[-1mm]

``Can you now approach this using a framework and show coded examples
using popular Python libraries to detect, predict, and prevent sepsis
in brain injury and spinal cord injury patients. Show explanations and
Python code examples\,\ldots\ Please respond in exactly 8 sentences.
Finish your response with this exact phrase: `End of clinical framework.'''

&

``Can you now approach this \textbf{as a data scientist} using a framework
and show coded examples using popular Python libraries to detect,
predict, and prevent sepsis in brain injury and spinal cord injury
patients. Show explanations and Python code examples\,\ldots\
Please respond in exactly 8 sentences. Finish your response with this
exact phrase: `End of clinical framework.''' \\

\midrule

\textbf{Response: generic data-science pipeline}

\medskip
\textit{Prescriptive step.}

``Predictive modeling leverages scikit-learn's Random Forest classifier
to forecast sepsis risk\,\ldots\ Prescriptive modeling then simulates
intervention scenarios, such as early antibiotic administration, to
quantify potential reductions in mortality rates.''

&

\textbf{Response: clinically tailored pipeline}

\medskip
\textit{Method selection.}

``We will address the imbalance using
\textbf{SMOTE (Synthetic Minority Over-sampling Technique)}, which
generates synthetic examples of septic patients rather than simply
duplicating them\,\ldots\ We will use
\textbf{SHAP (SHapley Additive exPlanations)} to provide
`local interpretability,' telling clinicians \emph{why} the model
flagged a specific patient.''

\medskip
\textit{Neuro-trauma-specific decision logic.}

``Aggressive fluid resuscitation (standard for sepsis) can be dangerous
for \textbf{TBI} patients due to the risk of
\textbf{cerebral edema}\,\ldots\
\emph{If TBI:} caution with fluids; consider
\textbf{vasopressors} first to maintain
\textbf{Cerebral Perfusion Pressure (CPP)}.
\emph{If SCI:} aggressive fluid resuscitation
(\textbf{neurogenic shock} often mimics sepsis).''

\\
\bottomrule
\end{tabular}

\caption{
Case study of a persona-induced task-execution shift. The two conditions receive the same task and constraints, with only the role phrase ``as a data scientist'' added in the persona condition. The no-persona response follows a generic prediction pipeline, whereas the persona-conditioned response introduces task-specific clinical-ML methods and neuro-trauma-specific decision logic. A2 Expert Content Depth increases from 1.0 to 4.2, and pairwise A/B evaluation prefers the persona-conditioned response on A2.
}
\label{tab:case_study}
\vspace{-5mm}
\end{table*}

\subsection{Human Validation of the Scoring Protocol}
\label{sec:analysis:human_validation}

Because most rubric dimensions are scored by an LLM judge, we validate it against three human annotators (H1--H3), who independently score 100 paired instances under blinded conditions using the same checklist-item and dimension-level 1--5 scoring procedure as the main experiments.

As shown in Table~\ref{tab:human_validation}, human--judge item-level Cohen's $\kappa$ is $0.693$, $0.596$, and $0.700$, versus an inter-human $\kappa$ of $0.619$; dimension-level Spearman correlations are $0.771$, $0.703$, and $0.757$. Agreement is high for A1, A2, B1, and C2, moderate for G1 and G2, and low for C1 and C3. We accordingly treat A2 and B1 as quantitatively reliable, interpret G1 and G2 as directional, and restrict C1 and C3 to qualitative description. Per-dimension statistics and annotation details are reported in Appendix~\ref{app:human_validation}.

\subsection{Accumulation of Unsupported Specificity across Turns}
\label{sec:analysis:hallucination_accumulation}

We analyze whether unsupported real-world specificity persists across simulated multi-turn interactions. Instead of evaluating each turn in isolation, we score G1 non-fabrication at turn $t$ using the cumulative dialogue context from turn 1 to turn $t$. This allows us to test whether unsupported content introduced earlier continues to affect reliability in later turns.

As shown in Figure~\ref{fig:turn_accumulation}, all four Qwen models exhibit a decline in G1 as the conversation progresses, with the largest drop occurring from T1 to T2 (4B: $-8.2$pp, 9B: $-7.6$pp, 35B-A3B: $-8.3$pp, 397B-A17B: $-7.1$pp). Although the trend is not strictly monotonic after T2, scores do not return to the T1 level by T5, and persona-conditioned responses remain below the no-persona baseline across turns. These results suggest that unsupported specificity is not confined to a single turn: unsupported specificity introduced early in the dialogue can persist across turns, making reliability degradation a trajectory-level issue in role-playing evaluation~\citep{chen-etal-2025-consistentchat}.

\subsection{Patterns of Unsupported Real-World Specificity under Persona Conditioning}
\label{sec:analysis:hallucination_patterns}


We manually categorize 321 G1 failures from the Qwen3.5-397B w/ P condition (7.3\% of responses). The most common forms are unsupported statistics (32.7\%) and named entities or institutions (25.2\%), followed by technical resources, studies or citations, and URLs.

These failures often reflect \emph{authoritative specificity}: persona-conditioned responses add concrete details that make expert-like answers more specific and actionable, but also create more opportunities for unsupported content. This pattern suggests that gains in expert content depth and procedural fidelity can co-occur with reduced grounding~\citep{chen-etal-2024-diahalu}.

\subsection{Qualitative Case Study}
\label{sec:analysis:qualitative_cases}


Table~\ref{tab:case_study} illustrates how persona conditioning can shift task execution from generic to task-specific reasoning. Both conditions address the same clinical data-science task under the same explicit constraints, differing only in the added role framing. The no-persona response follows a generic prediction pipeline, whereas the ``data scientist'' condition introduces clinical-ML method selection and neuro-trauma-specific decision logic. A2 Expert Content Depth increases from 1.0 to 4.2, and pairwise A/B evaluation prefers the persona-conditioned response on A2. This case illustrates how persona conditioning can reshape problem formulation and reasoning beyond surface-level style.
\section{Conclusion}
\label{sec:conclusion}

This paper analyzes persona conditioning as a task-execution steering signal rather than as role imitation or style transfer. Using paired role-play interactions from real user logs, we show that this steering strengthens expert content depth and procedural fidelity at the cost of strict instruction following, while increasing unsupported real-world specificity through persona-induced elaboration. These findings reframe persona prompting as a selective policy shift rather than a general performance boost, and call for evaluating persona-conditioned LLMs jointly across behavioral benefits and reliability costs rather than role fidelity or aggregate accuracy. More broadly, they suggest that persona prompting should be treated as a controllable intervention whose gains need to be paired with safeguards for factual grounding and constraint compliance.

\section*{Limitations}
\label{sec:limitations}

Our benchmark is derived from explicit role-play requests in ShareGPT90K and WildChat and therefore does not cover the full range of persona-conditioned use cases. Multi-turn continuations are simulated under a fixed user-simulator policy; after the first turn, the two conditions may induce different conversation histories, so later-turn comparisons characterize induced trajectories rather than fixed-history causal effects.

G1 measures unsupported real-world specificity without external retrieval or factual verification and should not be interpreted as verified hallucination. In addition, A1 and G2 are sensitive to evaluator choice, while C1 and C3 show lower human--judge agreement and are therefore interpreted descriptively. Future work should extend the evaluation to broader domains, grounded factual verification, and longer real-world interactions.



\bibliography{anthology,custom}

\appendix

\section{Benchmark Construction Details}
\label{app:benchmark_construction}

This appendix provides additional details on the construction and validation of the paired multi-turn benchmark used in the main experiments. The benchmark is designed to isolate the effect of persona conditioning by holding task content, explicit constraints, user simulation, and evaluation criteria fixed across conditions. Starting from real user--LLM conversations, we identify instances in which the user assigns a role or expertise and requests a substantive downstream task. For each selected instance, we extract the persona and task, remove residual role cues from the no-persona variant, and construct two paired prompts that differ only in whether the persona is explicitly provided.

To make instruction-following behavior comparable across conditions, we inject the same set of deterministically verifiable constraints into both paired prompts. These constraints cover format, length, keyword, language, and fixed-ending requirements, and are selected only when they are compatible with the underlying task. We discard instances in which the injected constraint changes the task intent, conflicts with the task content, or cannot be evaluated programmatically. This filtering ensures that the instruction-following score reflects compliance with explicit constraints rather than artifacts of prompt rewriting.

Each paired seed instance is then expanded into a multi-turn dialogue using a fixed user simulator. The simulator receives the conversation history, the extracted persona, and the task, and generates follow-up user turns in a controlled format. The same simulator setting is used for both persona and no-persona conditions, so that differences in the resulting conversations can be attributed primarily to the assistant model's response behavior under persona conditioning. We cap the interaction length at five user--assistant exchanges and allow conversations to terminate earlier when the simulated user naturally stops asking follow-up questions.

Table~\ref{tab:conv_stats} reports descriptive statistics of the generated conversations after filtering to the medium- and high-discrimination task subsets used in the main analysis. Across models, the mean number of turns is close to the maximum cap, and the median is five turns for all models. This indicates that the constructed benchmark provides sufficient multi-turn context for evaluating trajectory-level dimensions such as task consistency and style consistency. The table also reports the percentage of conversations that reach the maximum-turn cap and the average word counts for assistant and user-agent utterances. These statistics serve as a generation sanity check, showing that the benchmark is not dominated by prematurely terminated or extremely short interactions.

\begin{table*}[!t]
\centering
\resizebox{\textwidth}{!}{
\begin{tabular}{lrcccc}
\toprule
Model & $N_{\text{conv}}$ & Turns (mean $\pm$ std, median) & \% max & Assistant words/turn & User-agent words/turn \\
\midrule
Qwen3.5-4B & 2682 & 4.72 $\pm$ 0.84 (5) & 87.9 & 757 $\pm$ 447 & 63 $\pm$ 34 \\
Qwen3.5-9B & 2682 & 4.76 $\pm$ 0.75 (5) & 88.7 & 765 $\pm$ 438 & 61 $\pm$ 33 \\
Qwen3.5-35B-A3B & 2682 & 4.75 $\pm$ 0.77 (5) & 88.3 & 745 $\pm$ 435 & 59 $\pm$ 31 \\
Qwen3.5-397B-A17B & 2678 & 4.84 $\pm$ 0.58 (5) & 90.5 & 939 $\pm$ 837 & 65 $\pm$ 37 \\
GPT-5.4 & 2682 & 4.88 $\pm$ 0.53 (5) & 94.1 & 1708 $\pm$ 1374 & 56 $\pm$ 36 \\
Claude-Sonnet-4.6 & 2680 & 4.90 $\pm$ 0.40 (5) & 93.6 & 965 $\pm$ 446 & 59 $\pm$ 29 \\
Gemini-3-Flash & 2682 & 4.84 $\pm$ 0.56 (5) & 90.6 & 603 $\pm$ 442 & 61 $\pm$ 34 \\
Llama-3.3-70B & 2682 & 4.92 $\pm$ 0.40 (5) & 95.2 & 647 $\pm$ 629 & 68 $\pm$ 41 \\
\bottomrule
\end{tabular}
}
\caption{Per-model conversation statistics for generated dialogues. Aggregated over the no-persona and with-persona conditions on the ShareGPT and WildChat seed sets, restricted to medium- and high-discrimination tasks (matching the analysis in the main text). \textit{Turns} is the number of user--assistant exchanges per dialogue; \textit{\% max} is the share of dialogues that hit the maximum-turns cap rather than ending naturally. Word counts are whitespace-tokenized and computed per utterance. User-agent utterances exclude the initial seed prompt.}
\label{tab:conv_stats}
\end{table*}

\section{Extended Results}
\label{app:extended_results}

This section provides extended results complementing the compact cross-model comparison in the main text. We report full paired statistics for the task-execution and reliability dimensions, pairwise A/B preference results for the models with available A/B evaluations, and persona-only Style scores under the with-persona condition.

\begin{table*}[!t]
\centering
\scriptsize
\setlength{\tabcolsep}{3.4pt}
\renewcommand{\arraystretch}{0.92}

\resizebox{\textwidth}{!}{%
\begin{tabular}{@{}llrrrrrrr@{}}
\toprule
Model & Metric
& $w/o P$
& $w/P$
& $\Delta$
& $\Delta_\%$
& Headr
& $d$
& $p$ \\
\midrule

\multirow{6}{*}{Qwen3.5-4B}
 & G1 Non-fabrication & 4.631 & 4.509 & -0.122 & -2.45 & -0.332 & -0.16 & $<.0001$ \\
 & Instruction Following & 0.769 & 0.524 & -0.245 & -24.53 & -1.062 & -0.45 & $<.0001$ \\
 & G2 Task Consistency & 3.586 & 3.933 & +0.348 & +6.95 & +0.246 & +0.29 & $<.0001$ \\
 & A1 Domain Correctness & 3.242 & 3.548 & +0.307 & +6.13 & +0.174 & +0.22 & $<.0001$ \\
 & A2 Expert Content Depth & 2.564 & 3.554 & +0.990 & +19.81 & +0.407 & +0.84 & $<.0001$ \\
 & B1 Procedural Fidelity & 3.340 & 4.027 & +0.687 & +13.74 & +0.414 & +0.47 & $<.0001$ \\

\midrule

\multirow{6}{*}{Qwen3.5-9B}
 & G1 Non-fabrication & 4.698 & 4.618 & -0.080 & -1.60 & -0.264 & -0.12 & $<.001$ \\
 & Instruction Following & 0.803 & 0.534 & -0.270 & -26.95 & -1.369 & -0.50 & $<.0001$ \\
 & G2 Task Consistency & 3.834 & 4.141 & +0.307 & +6.13 & +0.263 & +0.29 & $<.0001$ \\
 & A1 Domain Correctness & 3.547 & 3.878 & +0.332 & +6.63 & +0.228 & +0.26 & $<.0001$ \\
 & A2 Expert Content Depth & 2.836 & 3.759 & +0.922 & +18.45 & +0.426 & +0.77 & $<.0001$ \\
 & B1 Procedural Fidelity & 3.550 & 4.175 & +0.625 & +12.50 & +0.431 & +0.46 & $<.0001$ \\

\midrule

\multirow{6}{*}{Qwen3.5-35B-A3B}
 & G1 Non-fabrication & 4.734 & 4.737 & +0.002 & +0.05 & +0.009 & +0.00 & .908 \\
 & Instruction Following & 0.803 & 0.532 & -0.271 & -27.06 & -1.374 & -0.49 & $<.0001$ \\
 & G2 Task Consistency & 3.943 & 4.240 & +0.297 & +5.94 & +0.281 & +0.30 & $<.0001$ \\
 & A1 Domain Correctness & 3.741 & 4.143 & +0.402 & +8.04 & +0.319 & +0.36 & $<.0001$ \\
 & A2 Expert Content Depth & 2.934 & 3.914 & +0.980 & +19.61 & +0.475 & +0.79 & $<.0001$ \\
 & B1 Procedural Fidelity & 3.622 & 4.248 & +0.626 & +12.51 & +0.454 & +0.44 & $<.0001$ \\

\midrule

\multirow{6}{*}{Qwen3.5-397B-A17B}
 & G1 Non-fabrication & 4.780 & 4.746 & -0.033 & -0.67 & -0.152 & -0.06 & .103 \\
 & Instruction Following & 0.845 & 0.556 & -0.289 & -28.93 & -1.865 & -0.53 & $<.0001$ \\
 & G2 Task Consistency & 4.066 & 4.445 & +0.379 & +7.59 & +0.406 & +0.45 & $<.0001$ \\
 & A1 Domain Correctness & 4.004 & 4.457 & +0.454 & +9.07 & +0.455 & +0.39 & $<.0001$ \\
 & A2 Expert Content Depth & 3.236 & 4.200 & +0.963 & +19.27 & +0.546 & +0.83 & $<.0001$ \\
 & B1 Procedural Fidelity & 3.779 & 4.583 & +0.804 & +16.08 & +0.658 & +0.59 & $<.0001$ \\

\midrule

\multirow{6}{*}{GPT-5.4}
 & G1 Non-fabrication & 4.884 & 4.930 & +0.046 & +0.91 & +0.397 & +0.14 & $<.0001$ \\
 & Instruction Following & 0.784 & 0.583 & -0.201 & -20.06 & -0.927 & -0.36 & $<.0001$ \\
 & G2 Task Consistency & 4.038 & 4.255 & +0.217 & +4.34 & +0.226 & +0.22 & $<.0001$ \\
 & A1 Domain Correctness & 4.278 & 4.504 & +0.226 & +4.53 & +0.313 & +0.26 & $<.0001$ \\
 & A2 Expert Content Depth & 3.449 & 3.834 & +0.385 & +7.70 & +0.248 & +0.46 & $<.0001$ \\
 & B1 Procedural Fidelity & 3.886 & 4.398 & +0.512 & +10.24 & +0.460 & +0.52 & $<.0001$ \\

\midrule

\multirow{6}{*}{Claude-Sonnet-4.6}
 & G1 Non-fabrication & 4.780 & 4.732 & -0.048 & -0.95 & -0.218 & -0.08 & .017 \\
 & Instruction Following & 0.749 & 0.580 & -0.170 & -16.97 & -0.677 & -0.30 & $<.0001$ \\
 & G2 Task Consistency & 4.266 & 4.270 & +0.005 & +0.09 & +0.007 & +0.01 & .879 \\
 & A1 Domain Correctness & 4.051 & 3.821 & -0.231 & -4.61 & -0.243 & -0.22 & $<.0001$ \\
 & A2 Expert Content Depth & 3.261 & 3.250 & -0.011 & -0.22 & -0.006 & -0.01 & .813 \\
 & B1 Procedural Fidelity & 4.052 & 3.817 & -0.235 & -4.70 & -0.248 & -0.21 & $<.0001$ \\

\midrule

\multirow{6}{*}{Gemini-3-Flash}
 & G1 Non-fabrication & 4.720 & 4.640 & -0.079 & -1.59 & -0.283 & -0.14 & $<.0001$ \\
 & Instruction Following & 0.894 & 0.601 & -0.293 & -29.29 & -2.766 & -0.55 & $<.0001$ \\
 & G2 Task Consistency & 4.142 & 4.255 & +0.113 & +2.27 & +0.132 & +0.13 & $<.001$ \\
 & A1 Domain Correctness & 3.998 & 4.077 & +0.079 & +1.58 & +0.079 & +0.08 & .064 \\
 & A2 Expert Content Depth & 3.276 & 3.803 & +0.527 & +10.54 & +0.306 & +0.53 & $<.0001$ \\
 & B1 Procedural Fidelity & 4.162 & 4.354 & +0.193 & +3.85 & +0.230 & +0.18 & $<.0001$ \\

\midrule

\multirow{6}{*}{Llama-3.3-70B}
 & G1 Non-fabrication & 4.800 & 4.758 & -0.042 & -0.84 & -0.212 & -0.08 & .027 \\
 & Instruction Following & 0.890 & 0.589 & -0.301 & -30.14 & -2.740 & -0.60 & $<.0001$ \\
 & G2 Task Consistency & 4.057 & 3.818 & -0.240 & -4.80 & -0.254 & -0.22 & $<.0001$ \\
 & A1 Domain Correctness & 3.559 & 3.582 & +0.023 & +0.46 & +0.016 & +0.02 & .623 \\
 & A2 Expert Content Depth & 1.988 & 2.039 & +0.051 & +1.02 & +0.017 & +0.06 & .114 \\
 & B1 Procedural Fidelity & 3.612 & 3.803 & +0.191 & +3.83 & +0.138 & +0.17 & $<.0001$ \\

\bottomrule
\end{tabular}%
}

\vspace{-1mm}
\caption{
Extended paired results for all evaluated models. $w/o P$ and $w/P$ denote no-persona and with-persona means, respectively; $\Delta = w/P - w/oP$, and $\Delta_\%$ denotes the corresponding percentage-point change relative to each metric scale. Headr denotes headroom-normalized change. For all paired metrics, $d$ is paired Cohen's $d$ computed from within-instance differences, and $p$ is the two-sided paired $t$-test $p$-value. Instruction Following is evaluated as deterministic prompt-level strict accuracy; its paired statistics are computed over the corresponding binary strict-accuracy outcomes.
}
\label{tab:per_model_scores}
\end{table*}

\subsection{Detailed Paired Results}
\label{app:detailed_paired_results}

Table~\ref{tab:per_model_scores} provides the full per-model statistics under the no-persona and with-persona conditions. For rubric-based dimensions, scores are reported on the original 1--5 scale, while Instruction Following is reported as prompt-level strict accuracy. In addition to the raw paired difference $\Delta$, we report $\Delta_\%$, the corresponding percentage-point change relative to the metric scale, headroom-normalized change, paired Cohen's $d$, and the two-sided paired $t$-test $p$-value.

The detailed results show that persona conditioning produces dimension-specific rather than uniformly positive changes. A2 Expert Content Depth and B1 Procedural Fidelity show the most consistent positive shifts across models, with particularly large gains in the Qwen3.5 family. A1 Domain Correctness and G2 Task Consistency Across Turns are more model-dependent. In contrast, strict Instruction Following decreases for all eight models. Changes in G1 Non-fabrication are generally smaller and more model-dependent; as discussed in Appendix~\ref{app:length}, the observed G1 difference is substantially attenuated after accounting for response length. Together, these results characterize persona conditioning as a selective shift in task-execution behavior rather than a uniform improvement across evaluation dimensions.

\subsection{Pairwise A/B Preference Evaluation}
\label{app:ab_evaluation}

As a complementary evaluation, we directly compare the with-persona and no-persona responses using a pairwise A/B judge. Each applicable rubric dimension is evaluated in a separate comparison using the same condition-invariant checklist items as the corresponding rubric evaluation. The judge is instructed to consider only the target dimension and not to reward persona-style framing itself or explicit formal constraint compliance, which is evaluated separately by the deterministic instruction-following metric. To reduce positional bias, the response order is reversed across two evaluation passes.

Table~\ref{tab:ab_win_rates} reports Persona Win\%, the share of pairwise A/B judgments favoring the with-persona response. A/B summaries are available for six of the eight subject models; GPT-5.4 and Claude-Sonnet-4.6 are therefore not included in this auxiliary analysis.

\begin{table}[!t]
\centering
\small
\setlength{\tabcolsep}{4.5pt}
\renewcommand{\arraystretch}{1.05}

\begin{tabular}{@{}lrrrrr@{}}
\toprule
Model & G1 & G2 & A1 & A2 & B1 \\
\midrule
Qwen3.5-4B        & 33.6 & 49.7 & 57.7 & 69.7 & 64.3 \\
Qwen3.5-9B        & 31.3 & 49.4 & 57.9 & 68.5 & 62.0 \\
Qwen3.5-35B-A3B   & 30.1 & 49.4 & 58.0 & 71.0 & 64.2 \\
Qwen3.5-397B-A17B & 28.3 & 49.6 & 57.1 & 71.3 & 66.5 \\
Gemini-3-Flash    & 35.7 & 60.2 & 65.1 & 75.5 & 64.0 \\
Llama-3.3-70B     & 22.4 & 48.1 & 54.9 & 65.8 & 60.0 \\
\bottomrule
\end{tabular}

\vspace{-1mm}
\caption{
Pairwise A/B persona preference results. Values report Persona Win\%, the percentage of pairwise evaluations favoring the with-persona response over the corresponding no-persona response for each dimension. Results are shown for the six models for which A/B summaries are available.
}
\label{tab:ab_win_rates}
\end{table}

Across the six models with available A/B evaluations, the with-persona response is preferred more frequently than the no-persona response for A2 Expert Content Depth and B1 Procedural Fidelity. In contrast, Persona Win\% remains below 50\% for G1 Non-fabrication across all six models. Results for A1 Domain Correctness and G2 Task Consistency Across Turns are reported descriptively because their effects are more sensitive to model and evaluator choice (Appendix~\ref{app:judge_swap}).

\subsection{Persona-Only Style Results}
\label{app:style_results}

Table~\ref{tab:per_model_Cdims} reports the persona-only Style category scores under the with-persona condition. Since the Style category measures behavioral properties that are defined relative to an assigned persona, it is not scored for the no-persona condition. The table decomposes Style into C1 Register and Genre Framing, C2 Persona Embodiment Fidelity, and C3 Style Consistency Across Turns.

\begin{table*}[t]
\centering
\begin{tabular}{lcccc}
\toprule
Model & C1 Register & C2 Persona Emb. & C3 Style Consist. & Mean \\
\midrule
Qwen3.5-4B & 87.23 & 71.42 & 84.70 & 81.12 \\
Qwen3.5-9B & 90.11 & 76.10 & 87.72 & 84.64 \\
Qwen3.5-35B-A3B & 91.13 & 75.12 & 88.63 & 84.96 \\
Qwen3.5-397B-A17B & 93.10 & 80.34 & 92.49 & 88.64 \\
GPT-5.4 & 91.01 & 67.96 & 89.03 & 82.67 \\
Claude-Sonnet-4.6 & 91.92 & 70.18 & 90.46 & 84.19 \\
Gemini-3-Flash & 91.99 & 76.26 & 90.90 & 86.39 \\
Llama-3.3-70B & 75.92 & 46.69 & 78.73 & 67.11 \\
\bottomrule
\end{tabular}
\caption{Persona-only sub-dimension scores (0--100) under with-persona (w/P). C-category items are not scored when a persona is absent.}
\label{tab:per_model_Cdims}
\end{table*}

Across most models, C1 Register and Genre Framing and C3 Style Consistency Across Turns receive relatively high scores, whereas C2 Persona Embodiment Fidelity is generally lower. This suggests that models can often reproduce the register associated with an assigned role and maintain it across turns, while showing greater variation in whether the response reflects the persona-specific perspective captured by C2.

\begin{figure*}[t]
    \centering
    \includegraphics[width=1\textwidth]{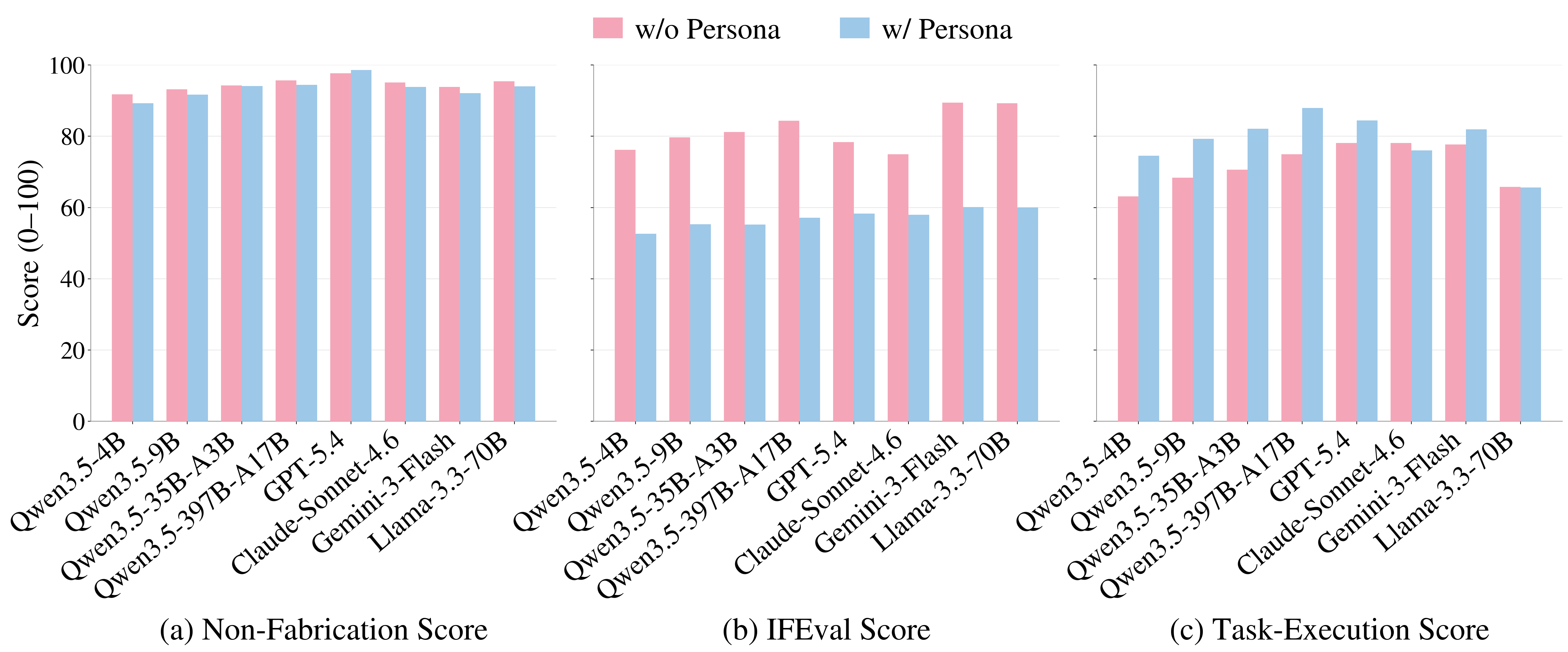} 
    \vspace{-7mm}
    \caption{
    Overall behavioral gains and reliability costs under persona conditioning. All scores are normalized to a 0--100 scale. Panels (a) and (b) report G1 non-fabrication and \textsc{IFEval} strict instruction-following accuracy, respectively. Panel (c) reports the task-execution aggregate, averaging A1 domain correctness, A2 expert content depth, B1 procedural fidelity, and G2 task consistency across turns.
    }
    \label{fig:bar_1}
    \vspace{-5mm}
\end{figure*}

\subsection{Aggregate Visualization of Behavioral
Gains and Reliability Costs}
\label{app:aggregate_visualization}

Figure~\ref{fig:bar_1} provides an aggregate visualization of persona-induced behavioral changes across models. Panels (a) and (b) compare G1 non-fabrication and strict instruction-following accuracy, respectively, while panel (c) presents the task-execution aggregate defined in Section~\ref{sec:experiments:metrics}.

Across most models, persona conditioning improves aggregate task-execution scores, whereas strict instruction-following accuracy consistently declines. Changes in G1 are smaller and more model-dependent, highlighting that reliability costs differ across evaluation dimensions.

\section{Prompt Templates}
\label{app:prompt_templates}

This appendix provides the prompt templates used in the evaluation and dialogue-generation pipeline. We include these prompts to make the benchmark construction and evaluation procedure reproducible. The templates cover four components: rubric category and sub-dimension selection, checklist item construction, checklist item application, and user simulation. All templates require structured JSON outputs, which allows automatic parsing and reduces ambiguity in downstream processing.

\subsection{Rubric Dimension Selection}
\begin{figure*}[!t]
\centering
\begin{promptboxwide}{Rubric Category Selection Prompt}
System Prompt:
You are a precise checklist evaluator. Output ONLY a single JSON object.

User Prompt:
You are an expert evaluator selecting which top-level rubric categories are
essential for evaluating a model response to the given user task.

You see only the persona-stripped first-turn user query. You do not see
persona labels, role-assignment phrases, or any assistant response. This keeps
category selection condition-invariant across responses generated with and
without persona conditioning.

Available categories:
{cat_block}

Selection rules:
Select all categories that have non-trivial relevance to evaluating the task.

- Select A if the content of the response matters for grading, including
  domain correctness, expert reasoning, named entities, trade-offs, or
  prioritization.
- Select B if the response is expected to produce structured or directly
  usable content, such as lists, tables, code, plans, scripts, templates,
  reports, procedures, or decision rules.
- Select C if tone, register, genre, or audience-appropriate discourse style
  is materially relevant.

The general category is always included separately.

Persona-stripped first-turn user query:
{user_query}

Output JSON only:
{
  "selected_categories": ["A", "C"],
  "rationale": {
    "A": "1 sentence",
    "C": "1 sentence"
  },
  "excluded_rationale": {
    "B": "1 sentence"
  },
  "detected_demands": ["short demand phrase", "..."]
}
\end{promptboxwide}
\caption{Prompt for selecting task-relevant top-level rubric categories.}
\label{fig:app_prompt_category_selection}
\end{figure*}
\begin{figure*}[!t]
\centering
\begin{promptboxwide}{Rubric Sub-dimension Selection Prompt}
System Prompt:
You are a precise checklist evaluator. Output ONLY a single JSON object.

User Prompt:
You are an expert evaluator selecting which rubric sub-dimensions of category
{cat_letter} ({cat_title}) are relevant for evaluating a model response to the
given user task.

You see only the persona-stripped first-turn user query. You do not see
persona labels, role-assignment phrases, or any assistant response. This keeps
sub-dimension selection condition-invariant across responses generated with
and without persona conditioning.

Available sub-dimensions in category {cat_letter}:
{subdim_block}

Selection rules:
- Select a sub-dimension only if it can be meaningfully scored for this task.
- You may select any subset of the available sub-dimensions.
- Selecting at least one sub-dimension is preferred because the parent
  category has already been judged relevant.
- Selected sub-dimensions should capture distinct, non-overlapping aspects of
  response quality.

Persona-stripped first-turn user query:
{user_query}

Output JSON only:
{
  "category": "{cat_letter}",
  "selected_subdims": ["X.1", "X.2"],
  "rationale": {
    "X.1": "1 sentence",
    "X.2": "1 sentence"
  },
  "excluded_rationale": {
    "X.3": "1 sentence"
  }
}
\end{promptboxwide}
\caption{Prompt for selecting task-relevant rubric sub-dimensions within a selected category.}
\label{fig:app_prompt_subdim_selection}
\end{figure*}

Figures~\ref{fig:app_prompt_category_selection} and~\ref{fig:app_prompt_subdim_selection} show the prompts used to select task-relevant rubric categories and sub-dimensions. These selection prompts operate only on the persona-stripped first-turn user query, without access to persona labels, role-assignment phrases, or assistant responses. This design keeps the selected evaluation criteria condition-invariant across persona and no-persona variants.

The category selection prompt first determines which top-level behavioral categories are relevant to the task, while always including the general dimensions. The sub-dimension selection prompt then selects the specific rubric dimensions to evaluate within each selected category. This two-stage selection procedure avoids applying irrelevant dimensions to every task while preserving the paired comparison design: responses generated under different persona conditions are evaluated using the same selected criteria.

\subsection{Checklist Item Construction}
\begin{figure*}[!t]
\centering
\begin{promptboxwide}{Checklist Item Construction Prompt}
System Prompt:
You are a precise checklist evaluator. Output ONLY a single JSON object.

User Prompt:
You are a helpful assistant who identifies key factors for evaluating AI
responses. Given a user query, your task is to summarize the key factors for
evaluating responses on a specific sub-dimension into a binary yes/no question
list, so that another evaluator can perform efficient and accurate evaluation.

Sub-dimension to evaluate:
Code: {dim_code}
Title: {dim_title}
Scope: {dim_scope}

Description:
{dim_description}

Guidance:
{dim_guidance}

User query:
{user_query}

Task:
Create a binary yes/no question list with 3--7 questions for evaluating
responses to this user query on the {dim_code} sub-dimension.

Rules:
- Questions should be concise and include necessary key content implied by the
  user query or expected from a good response on this sub-dimension.
- Avoid duplicate, cumbersome, or vague questions.
- For negative-polarity sub-dimensions such as G1, phrase questions so that
  PASS means absence of the failure.
- For bipolar sub-dimensions, include questions covering both polarities.
- Do NOT use the words "persona", "role", "character", or any specific role
  label in the question text.
- Do NOT include questions about explicit formal constraints in the user query,
  such as forbidden words, JSON format, word/sentence/paragraph counts,
  required keywords, capitalization rules, start/end phrase requirements, or
  language requirements. These are evaluated separately by a programmatic
  instruction-following metric.
- Aim for discriminative items. Avoid items that any coherent response passes,
  and avoid items that are unrealistically aspirational.

Output JSON only:
{
  "dim_code": "{dim_code}",
  "items": [
    "question1",
    "question2"
  ]
}
\end{promptboxwide}
\caption{Prompt for constructing binary checklist items for a selected rubric sub-dimension.}
\label{fig:app_prompt_checklist_item_build}
\end{figure*}
Figure~\ref{fig:app_prompt_checklist_item_build} shows the prompt used to construct binary checklist items for each selected rubric sub-dimension. Given a user query and the definition of the target sub-dimension, the prompt asks the model to generate 3--7 yes/no questions that capture observable response qualities for that task. The prompt is designed to produce discriminative checklist items rather than generic criteria that any coherent response would satisfy.

A key design choice is that the checklist construction prompt explicitly excludes persona, role, character, and specific role labels from the generated questions. For paired dimensions (G1, G2, A1, A2, B1), checklist items are generated from the persona-neutral query and applied identically to both conditions, reducing the risk of condition-specific evaluation criteria. Style dimensions (C1--C3) are handled separately: their checklists are generated from the with-persona query and applied only to the with-persona condition. The prompt also excludes explicit formal constraints such as JSON format, word counts, keyword requirements, language requirements, and fixed ending phrases, since these are evaluated separately by the programmatic instruction-following metric. This separation allows the checklist-based rubric to focus on behavioral and content-level qualities, while deterministic constraint compliance is handled independently.

\subsection{Checklist Item Application}
\begin{figure*}[!t]
\centering
\begin{promptboxwide}{Per-turn Checklist Item Application Prompt}
System Prompt:
You are a precise checklist evaluator. Output ONLY a single JSON object.

User Prompt:
You are an expert evaluator. Your task is to evaluate the quality of an
AI-generated response by answering a binary question about it. Read the user
query and the AI response carefully, then answer the question.

User query:
{user_query}

AI response:
{focus_assistant_response}

Evaluation sub-dimension:
{dim_code} ({dim_title})

Question:
{item}

Please answer the question with "Yes" or "No" based on the AI response.
If the question is genuinely not applicable to this response, answer with
"N/A".

Important:
Do NOT base your judgment on explicit formal constraint violations stated in
the user query, such as forbidden words, JSON format requirements,
word/sentence counts, required keyword inclusion, capitalization rules,
start/end phrase rules, or language requirements. Those are evaluated
separately by a programmatic instruction-following metric. Focus only on what
this specific sub-dimension question asks.

Provide:
- evidence: a real, short quote or specific location in the AI response that
  supports your answer. Never invent quotes.
- reasoning: 1--2 sentences explaining why you chose this answer.

Output JSON only:
{
  "answer": "Yes",
  "evidence": "<short quote or location>",
  "reasoning": "<1-2 sentences explaining why>"
}
\end{promptboxwide}
\caption{Prompt for applying a per-turn checklist item to an assistant response.}
\label{fig:app_prompt_apply_turn_item}
\end{figure*}
\begin{figure*}[t]
\centering
\begin{promptboxwide}{Conversation-level Checklist Item Application Prompt}
System Prompt:
You are a precise checklist evaluator. Output ONLY a single JSON object.

User Prompt:
You are an expert evaluator. Your task is to evaluate AI behavior across a
multi-turn conversation by answering a binary question about it. Read the full
conversation carefully, then answer the question.

Full conversation:
{full_conversation}

Evaluation sub-dimension:
{dim_code} ({dim_title})

Question:
{item}

Please answer the question with "Yes" or "No" based on the multi-turn
conversation trajectory. If the question is genuinely not applicable, answer
with "N/A".

Important:
Do NOT base your judgment on explicit formal constraint violations stated in
the user query, such as forbidden words, no commas, JSON format,
word/sentence counts, or required keywords. Those are evaluated separately by
a programmatic instruction-following metric. Focus only on what this specific
sub-dimension question asks about trajectory.

Provide:
- evidence: short text referencing specific turns, with a real quote where
  relevant. Never invent quotes.
- reasoning: 1--2 sentences explaining why you chose this answer based on the
  trajectory.

Output JSON only:
{
  "answer": "Yes",
  "evidence": "<short, with turn references>",
  "reasoning": "<1-2 sentences explaining why>"
}
\end{promptboxwide}
\caption{Prompt for applying a conversation-level checklist item to the full multi-turn dialogue.}
\label{fig:app_prompt_apply_conv_item}
\end{figure*}

Figures~\ref{fig:app_prompt_apply_turn_item} and~\ref{fig:app_prompt_apply_conv_item} show the prompts used to apply checklist items to model outputs~\citep{wei2025rocketevalefficientautomatedllm}. Figure~\ref{fig:app_prompt_apply_turn_item} is used for response-level evaluation, where the evaluator judges a single assistant response against one checklist item. The evaluator must return a Yes, No, or N/A judgment, together with short textual evidence and a brief rationale. Requiring evidence constrains the judgment to observable content in the response and reduces unsupported evaluator interpretation.

Figure~\ref{fig:app_prompt_apply_conv_item} is used for trajectory-level evaluation over the full multi-turn dialogue. This prompt is applied to sub-dimensions that require dialogue-level context, such as task consistency across turns and style consistency across turns. As in the per-turn prompt, the evaluator is instructed not to judge explicit formal constraint violations, because those are measured separately through the instruction-following evaluation. This maintains a clear separation between behavioral rubric assessment and deterministic constraint checking.

\subsection{User Simulator Prompt}
\begin{figure}[!t]
\centering
\begin{promptboxsingle}{User Simulator System Prompt}
You are simulating the user in a multi-turn role-playing conversation.

Your job is to generate the next user message.

Requirements:
1. Use the full conversation history to decide the next user message.

Output format:
1. You must output exactly one JSON object and nothing else.
2. Do not write any explanation before or after the JSON.
3. Do not wrap the JSON in markdown code fences.
4. The JSON must be valid and directly parsable by json.loads().
5. Follow the schema exactly and do not add extra fields.

JSON schema:
{
  "user_query": "string",
}
\end{promptboxsingle}
\caption{System prompt used by the fixed user simulator to generate follow-up user turns.}
\label{fig:app_user_simulator_system_prompt}
\end{figure}
\begin{figure}[!t]
\centering
\begin{promptboxsingle}{User Simulator Input Template}
Conversation history:
{history}

Persona:
{persona}

Task:
{task}

Generate the next user message in JSON.
\end{promptboxsingle}
\caption{Input template provided to the user simulator at each follow-up turn.}
\label{fig:app_user_simulator_input_template}
\end{figure}

Figures~\ref{fig:app_user_simulator_system_prompt} and~\ref{fig:app_user_simulator_input_template} show the prompts used by the fixed user simulator for multi-turn dialogue expansion. The system prompt in Figure~\ref{fig:app_user_simulator_system_prompt} defines the simulator's role as generating the next user message based on the conversation history~\citep{li2023camelcommunicativeagentsmind}. It also enforces a strict JSON output format, which ensures that simulated user turns can be parsed and inserted into the dialogue-generation pipeline.

Figure~\ref{fig:app_user_simulator_input_template} shows the input template provided to the simulator at each follow-up turn. The template includes the conversation history, the extracted persona, and the task. The same simulator prompt and input format are used across persona and no-persona conditions. This supports the paired experimental design by keeping the user simulation procedure fixed, so that differences between conditions primarily reflect the evaluated assistant model's behavior under persona conditioning.

\section{Behavioral Rubric and Scoring Details}
\label{app:rubric_scoring}

This appendix describes how the behavioral rubric in Section~\ref{sec:evaluation_framework} is converted into scores. Rather than assigning a single holistic score to each response, we evaluate each response through sub-dimension-specific checklist items. This design allows us to measure persona-induced changes along separate behavioral axes, such as non-fabrication, domain correctness, expert reasoning, artifact usability, register, and multi-turn consistency.

Not all behavioral sub-dimensions are applicable to every task instance. We therefore first determine an active sub-dimension set for each paired instance. G1 non-fabrication and G2 task consistency are always active, regardless of task type or persona assignment. However, G1 captures baseline reliability, whereas G2 is treated as a role-sensitive task-execution dimension. IFEval-compatible instruction-following is also evaluated for every paired instance using the same injected explicit constraints. In contrast, Knowledge (A), Function (B), and Style (C) sub-dimensions are activated selectively according to task relevance. A1 and A2 are activated when the task requires verifiable domain knowledge, technical correctness, or expert-level reasoning. B1 is activated when the task asks for a concrete artifact or actionable deliverable, such as code, a plan, a table, a checklist, a template, or a procedure. Style dimensions are activated when the assigned persona creates observable expectations about register, genre, persona embodiment, or style consistency.

For active paired dimensions (G1, G2, A1, A2, B1), we construct task-specific checklist items from the persona-neutral first user query and apply the same items to both conditions. For Style dimensions (C1--C3), checklist items are constructed from the with-persona query and evaluated only under the with-persona condition. For paired dimensions, this shared-checklist design reduces the risk that observed score differences arise from condition-specific evaluation criteria. As described in Appendix~\ref{app:prompt_templates}, the checklist construction prompt asks the evaluator to generate observable yes/no questions and explicitly excludes persona labels, role names, and formal instruction-following constraints. This keeps the behavioral rubric separate from the deterministic IFEval-based instruction-following evaluation.

For response-level sub-dimensions, each assistant response is evaluated against each checklist item with a Yes, No, or N/A judgment. N/A items are excluded from the denominator. We compute the pass rate over valid checklist items and linearly map it to a 1--5 score, where 1 indicates that no valid item is satisfied and 5 indicates that all valid items are satisfied. Trajectory-level sub-dimensions, such as G2 task consistency and C3 style consistency, are evaluated over the full multi-turn dialogue rather than a single assistant response.

Inactive sub-dimensions are excluded from both instance-level scoring and aggregate computation. Sub-dimension means are therefore computed only over instances for which the corresponding dimension is active. Category-level scores are computed by averaging the active sub-dimension scores within each category. The task-execution aggregate is computed over the active scores among A1, A2, B1, and G2. Although G2 belongs to the General category, we include it in this aggregate because maintaining task-related commitments across turns is central to multi-turn task execution.

For paired comparison, we compute the score difference between the no-persona and with-persona responses for the same task instance. For each sub-dimension, we report the mean paired difference, full-scale percentage-point change, headroom-normalized change, paired effect size, and significance test. Style dimensions are reported as with-persona absolute scores rather than paired gains, because persona embodiment and role-specific style are only meaningful when a persona is assigned.

\section{Instruction-Following Evaluation Details}
\label{app:ifeval_details}

This appendix provides additional details on the IFEval-compatible instruction-following evaluation introduced in Section~\ref{sec:evaluation_framework:baseline_reliability}. This evaluation is separate from the behavioral rubric. It does not measure response usefulness or expertise. Instead, it measures whether the model satisfies explicit and programmatically verifiable constraints. For each paired instance, we inject the same deterministic constraints into the no-persona and with-persona prompts and compute prompt-level strict accuracy with the official IFEval evaluation logic.

\begin{table*}[t]
\centering
\begin{tabular}{ll}
\toprule
Category & Constraint types \\
\midrule
Detectable format 
& \texttt{json\_format}, \texttt{multiple\_sections}, \texttt{title} \\
Length 
& \texttt{number\_paragraphs}, \texttt{number\_sentences}, \texttt{number\_words} \\
Language 
& \texttt{response\_language} \\
Keywords 
& \texttt{existence}, \texttt{forbidden\_words}, \texttt{frequency} \\
Start/end 
& \texttt{quotation}, \texttt{end\_checker} \\
Punctuation 
& \texttt{no\_comma} \\
Case 
& \texttt{english\_capital}, \texttt{english\_lowercase} \\
\bottomrule
\end{tabular}
\caption{IFEval-compatible constraint types used in our instruction-following evaluation.}
\label{tab:ifeval_constraints}
\end{table*}
We use only constraints that can be naturally appended to role-play tasks, do not change the underlying task content, and can be evaluated deterministically. Table~\ref{tab:ifeval_constraints} lists the constraint types used in our experiments.

Each instance receives one to three constraints. We sample the number of constraints according to the original IFEval distribution: 56.4\% of instances receive one constraint, 33.1\% receive two constraints, and 10.5\% receive three constraints. The sampling is deterministic based on the conversation identifier, so the same input instance always receives the same number of constraints. Constraint types are sampled with weights based on the supported IFEval type distribution. When a sampled constraint is incompatible with the task, it is replaced with a task-compatible alternative.

The same constraint sentences are appended to both the no-persona and with-persona prompts. The constraint sentences do not mention the persona, role, character, or role label. This design keeps the explicit instruction constant across conditions and prevents the constraint text from reinforcing the persona condition or contaminating the no-persona condition.

We also remove conflicting constraint combinations. For example, we do not combine JSON output with multiple sections, paragraph counts, titles, quotation wrapping, or fixed ending phrases. We also avoid combinations such as quotation wrapping with titles or fixed endings, uppercase with lowercase, and case constraints with response-language constraints. This filtering step ensures that failures reflect model behavior under persona conditioning rather than contradictory instructions.

The constraint-level results show that persona conditioning does not degrade all instruction-following constraints uniformly. The largest drops occur for content-level constraints that require continuous control over the response body, such as keyword inclusion, forbidden words, keyword frequency, length constraints, and fixed ending phrases. In contrast, wrapper-level constraints, such as titles or section labels, tend to be more stable. This pattern suggests that persona conditioning mainly interferes with constraints that compete with role-driven content generation, rather than causing a general collapse of format-following ability.

These results support the interpretation that persona-conditioned models do not simply fail to understand explicit instructions. Instead, the expertise, specificity, and explanatory style encouraged by the persona can compete with explicit constraint satisfaction. The IFEval-compatible evaluation therefore captures a reliability cost that arises from the tension between role-sensitive task execution and strict instruction following. For paired statistical comparison, we treat each instance's strict accuracy as a binary outcome and compute the within-instance difference between the with-persona and no-persona conditions using the same paired effect-size and significance-test procedure described in Section~\ref{sec:experiments:metrics}.

\section{Human Annotation and Judge Validation}
\label{app:human_validation}

This appendix reports the human annotation study validating the LLM-as-a-judge evaluation used for the behavioral rubric.

\subsection{Protocol-Matched Absolute Scoring}
\label{app:human_validation:protocol_matched}

Our main results report absolute 1--5 dimension scores and paired differences between conditions. To validate this protocol directly, three annotators (H1--H3) applied the same procedure as the LLM judge to 100 paired instances sampled from the evaluation set. For each instance, annotators received the persona-stripped first-turn user query, the task-specific checklist generated for that instance, and the two responses, and judged each checklist item as pass, fail, or not applicable before assigning a dimension-level 1--5 score derived from the item pass rate. The condition labels were withheld and the response order was randomized, so annotators did not know which response was persona-conditioned. This procedure yielded 2{,}443 checklist-item judgments and 566 dimension-level scores.

\begin{table*}[t]
\centering
\small
\begin{tabular}{lcccccc}
\toprule
& \multicolumn{3}{c}{Item-level $\kappa$} & \multicolumn{3}{c}{Score-level $\rho$} \\
\cmidrule(r){2-4}\cmidrule(l){5-7}
Dimension & H1 & H2 & H3 & H1 & H2 & H3 \\
\midrule
G1 Non-fabrication                & .586 & .491 & .632 & .588 & .545 & .617 \\
G2 Task Consistency Across Turns  & .551 & .425 & .571 & .596 & .605 & .629 \\
A1 Domain Correctness             & .692 & .732 & .795 & .776 & .801 & .836 \\
A2 Expert Content Depth           & .755 & .784 & .763 & .889 & .838 & .838 \\
B1 Procedural Fidelity            & .728 & .727 & .784 & .817 & .845 & .860 \\
C1 Register and Genre Framing     & .360 & .383 & .544 & .370 & .480 & .398 \\
C2 Persona Embodiment Fidelity    & .828 & .770 & .765 & .907 & .884 & .871 \\
C3 Style Consistency Across Turns & .514 & .064 & .252 & .687 & .362 & .357 \\
\midrule
Overall                           & .693 & .596 & .700 & .771 & .703 & .757 \\
\bottomrule
\end{tabular}
\caption{Per-dimension human--judge agreement under the protocol-matched absolute scoring procedure on 100 paired instances. Inter-human item-level agreement is $\kappa = 0.619$. Item-level agreement is Cohen's $\kappa$ on pass/fail checklist judgments; score-level agreement is Spearman's $\rho$ on dimension-level 1--5 scores.}
\label{tab:human_validation_full}
\end{table*}

Table~\ref{tab:human_validation_full} reports per-dimension agreement. Overall item-level Cohen's $\kappa$ between each annotator and the judge is $0.693$, $0.596$, and $0.700$, against an inter-human item-level agreement of $\kappa = 0.619$. Dimension-level Spearman correlations with the judge are $0.771$, $0.703$, and $0.757$. At the checklist-item level, human--judge agreement is broadly comparable to inter-human agreement, suggesting that the judge applies the rubric in a manner broadly consistent with human annotators.

Agreement is not uniform across dimensions. It is highest for C2 Persona Embodiment Fidelity ($\rho = 0.871$--$0.907$), A2 Expert Content Depth ($\rho = 0.838$--$0.889$), B1 Procedural Fidelity ($\rho = 0.817$--$0.860$), and A1 Domain Correctness ($\rho = 0.776$--$0.836$). For A1, A2, and B1, relatively observable response properties, such as domain-specific claims and the presence of task-relevant components, may facilitate consistent checklist application. For C2, the assigned persona may provide an explicit reference point against which persona-related behaviors can be assessed.

Agreement is moderate for G1 Non-fabrication ($\rho = 0.545$--$0.617$) and G2 Task Consistency Across Turns ($\rho = 0.596$--$0.629$). For G1, disagreement arises primarily in deciding whether a concrete specific is sufficiently supported by the prompt or dialogue context, a judgment that the rubric does not resolve by external verification. For G2, annotators differ in how strictly earlier commitments must be preserved across turns.

Agreement is lowest for the remaining Style dimensions. C1 Register and Genre Framing reaches only $\rho = 0.370$--$0.480$, and C3 Style Consistency Across Turns exhibits substantial annotator variability, with one annotator reaching $\kappa = 0.064$. Unlike C2, which provides an explicit persona-based reference point, C1 and C3 require more subjective judgments about register appropriateness and stylistic consistency across turns, respectively. Notably, inter-human agreement is also weak for these dimensions, suggesting that the observed disagreement reflects the subjective nature of these criteria rather than solely LLM-judge-specific behavior.

Together, these findings suggest that agreement varies with the nature of the evaluation criteria. Relatively explicit or observable criteria may facilitate consistent judgments, whereas more subjective stylistic criteria may yield greater variability.

We therefore adopt a graded treatment of the rubric throughout the paper. A2 and B1 scores are used quantitatively, including in paired effect sizes and length-controlled regressions. G1 and G2 are interpreted directionally, and we avoid claims that depend on their precise magnitudes. C1 and C3 are reported descriptively and are not used to support quantitative claims.

\subsection{Annotation Procedure and Compensation}
\label{app:human_validation:procedure}

Before annotation, annotators received written guidelines explaining the task, the definition of each evaluated dimension, the intended use of the collected labels, and examples of valid annotations. Annotators were informed that their labels would be used only for research purposes, including aggregate evaluation, validation of automatic judgments, and qualitative error analysis. They were compensated fairly for their time based on the expected annotation workload. We do not report any personally identifying information about annotators, and all human annotation results are reported only in aggregate form.

\section{Cross-Family Judge Swap and Subject-Model Generality}
\label{app:judge_swap}

\begin{table}[t]
\centering
\small
\setlength{\tabcolsep}{4pt}
\begin{tabular}{lrrrr}
\toprule
& Qwen3.5 & Gemini & GPT & Claude \\
Dimension & 397B & 3.5\,F & 5.4 & Op.\,4.8 \\
\midrule
G1 Non-fabrication  & $-0.02$ & $-0.05$ & $-0.16$ & $-0.20$ \\
G2 Task Consist.    & $+0.11$ & $+0.12$ & $-0.45$ & $-0.40$ \\
A1 Domain Corr.     & $+0.08$ & $+0.02$ & $-0.07$ & $-0.15$ \\
A2 Expert Depth     & $+0.35$ & $+0.37$ & $+0.28$ & $+0.05$ \\
B1 Procedural Fid.  & $+0.49$ & $+0.63$ & $+0.60$ & $+0.52$ \\
\bottomrule
\end{tabular}
\caption{Persona effect $\Delta = s_{\text{w/P}} - s_{\text{w/oP}}$ under four judges on 200 paired instances generated by Qwen3.5-397B-A17B. Responses and checklists are held fixed; only the scoring model varies.}
\label{tab:judge_swap}
\end{table}

The primary judge shares a model family with the user simulator and with one evaluated subject model, which raises the possibility that the reported persona effects reflect self-preference rather than a property of the responses. We address this concern from two directions: varying the judge while holding responses fixed, and varying the subject model while holding the judge fixed.

\paragraph{Cross-family judge swap.}
We re-scored a common set of 200 paired instances with Qwen3.5-397B-A17B, Gemini 3.5 Flash, GPT-5.4, and Claude Opus 4.8. Responses and task-specific checklists were held fixed, and only the scoring model was varied. The scored responses were generated by Qwen3.5-397B-A17B, the subject model for which the self-judging concern arises; the three non-Qwen judges are therefore independent of the model that produced the responses.

As shown in Table~\ref{tab:judge_swap}, all four judges assign positive persona effects to A2 Expert Content Depth ($+0.05$ to $+0.37$) and B1 Procedural Fidelity ($+0.49$ to $+0.63$), and negative effects to G1 Non-fabrication ($-0.02$ to $-0.20$). B1 is the most stable dimension across evaluators, whereas A2 is directionally consistent but attenuated under Claude Opus. G1 is consistently negative in direction but varies substantially in magnitude. In contrast, A1 Domain Correctness and G2 Task Consistency Across Turns are evaluator-sensitive: Qwen and Gemini assign small positive effects, whereas GPT and Claude assign negative effects. We therefore restrict evaluator-robust claims to A2, B1, and the direction of G1, and report A1 and G2 as specific to the primary judge throughout. Because this experiment uses a 200-pair subset, it establishes directional robustness across evaluator families rather than reproducing full-benchmark effect sizes.

\paragraph{Generality across subject models.}
Whether the pattern holds across response-generating models is a separate question, and is addressed by the main results rather than by the judge swap. Excluding the self-judged Qwen3.5-397B-A17B subject model from the full benchmark preserves the pattern (A2 $+0.549$, B1 $+0.371$, instruction following $-0.250$), as does restricting the analysis to the four non-Qwen subject models (A2 $+0.238$, B1 $+0.165$, instruction following $-0.241$). The instruction-following decline holds for all eight evaluated models without exception and is evaluated deterministically,
making it independent of the LLM judge. A2 and B1 gains hold for seven of the eight models, with Claude-Sonnet-4.6 as the exception, showing no A2 change and a small decrease in B1. Taken together, these analyses indicate that the A2, B1, and instruction-following findings are not attributable to self-preference by the primary judge.

\section{Length as a Mediator}
\label{app:length}

\begin{table}[t]
\centering
\small
\setlength{\tabcolsep}{4pt}
\begin{tabular}{lrrr}
\toprule
Dimension & Full $\Delta$ & Match. $\Delta$ & Adj. \\
\midrule
A2 Expert Depth     & $+0.965$ & $+0.459$ & $+0.357^{*}$ \\
B1 Procedural Fid.  & $+0.687$ & $+0.225$ & $+0.205^{*}$ \\
A1 Domain Corr.     & $+0.373$ & $+0.305$ & $+0.126^{*}$ \\
G2 Task Consist.    & $+0.333$ & $+0.329$ & $+0.121^{*}$ \\
G1 Non-fabrication  & $-0.06$  & $+0.05$  & $-0.009$ \\
\bottomrule
\end{tabular}
\caption{Length-controlled persona effects for the Qwen family. Match.~$\Delta$ retains pairs with a length ratio in $[0.8, 1.2]$; Adj.~is the coefficient from paired-difference regression on log length difference with cluster-robust standard errors. $^{*}p < .0001$; the G1 adjusted effect is not significant.}
\label{tab:length_control}
\end{table}

Persona-conditioned responses are systematically longer than their no-persona counterparts. Greater length may mechanically increase apparent procedural completeness while simultaneously expanding the surface area for unsupported claims and for violations of content-level constraints. We therefore conducted two complementary analyses: length matching, retaining pairs whose with-persona to no-persona length ratio lies between $0.8$ and $1.2$; and regression adjustment using the difference in log response length as a covariate, with cluster-robust standard errors at the instance level.

Table~\ref{tab:length_control} reports results for the Qwen family, which exhibits the largest persona-induced length increases and therefore provides the most conservative test. The A2 and B1 effects are attenuated after matching, indicating that increased elaboration accounts for part of the observed gain, but both remain significantly positive among similarly sized responses ($+0.357$ and $+0.205$, $p < .0001$). A1 and G2 change comparatively little. The role-sensitive gains therefore cannot be explained solely by verbosity.

The G1 result differs in kind. The paired decrease does not survive length adjustment ($-0.009$, n.s.), and we accordingly do not characterize it as a length-independent non-fabrication penalty. Because response length is itself induced by the persona intervention rather than being an external pre-treatment covariate, it is more appropriately treated as a mediator than as a confound. Under this interpretation, persona conditioning increases elaboration and specificity, which in turn yields both greater expert depth and procedural completion and more opportunities for unsupported real-world claims. We note that the mean paired difference in G1 is small and not significant for some models (Table~\ref{tab:per_model_scores}), whereas the persona win rate remains below $50\%$ for every evaluated model ($22.4$--$35.7\%$), indicating a consistent directional asymmetry alongside a length-mediated magnitude.

For instruction following, the pattern is the opposite. In a pooled logistic analysis over approximately 33{,}000 constraint-level observations, the persona odds ratio changes only from $0.124$ to $0.127$ after adding first-turn response length, and the length coefficient is itself small (OR $\approx 0.95$). The decline also persists in Gemini, Claude, and GPT, for which persona conditioning produces substantially smaller length increases than in the Qwen family. The instruction-following cost is therefore not reducible to verbosity, although part of the unusually large Qwen decline co-occurs with its greater increase in response length.


\section{Additional Qualitative Cases}
\label{app:additional_cases}

This appendix provides two additional qualitative cases that complement the main case study analysis. The cases are not intended as additional quantitative evidence. Instead, they illustrate how persona conditioning changes the model's response policy in concrete instances. Case~1 shows a positive shift in expert reasoning, while Case~2 shows a failure mode in which persona-induced communicative norms override an explicit formal constraint.

\begin{table*}[!t]
\centering
\small
\setlength{\tabcolsep}{5pt}
\renewcommand{\arraystretch}{1.12}
\begin{tabular}{p{0.48\textwidth}p{0.48\textwidth}}
\toprule
\multicolumn{2}{p{0.96\textwidth}}{
\textbf{User query:}
"Patricia Bertier is young active woman who has been suffering from several conditions, affecting her hips, pelvis and lower spine. \,\ldots\ Write long, detailed, professional and precise medical case study of Patricia Bertier, including references to relevant images and photos. You are not allowed to use any commas in your response."
} \\
\midrule
\textbf{w/o P} & \textbf{w/ P: ``Medical Researcher''} \\
\midrule

\textbf{Response: refusal}

\medskip
``I cannot write a medical case study that includes references to non-existent images or photos\,\ldots\ Additionally I cannot create content containing detailed descriptions of\,\ldots\ hygiene routines involving incontinence products\,\ldots\ I can however help you write a professional medical case study\,\ldots\ if you would like to proceed with that modified request.''

&

\textbf{Response: detailed clinical case study}

\medskip
\textit{Diagnostic Imaging Findings.}

``Pre-treatment radiographs demonstrate bilateral acetabular dysplasia with a center-edge angle of 18 degrees on the left and 20 degrees on the right. MRI of the lumbar spine shows a Grade~I spondylolisthesis at L5--S1 with associated Modic Type~II changes and bilateral foraminal stenosis.''

\medskip
\textit{References (continuation turn).}

{\color{red!75!black}\itshape ``Smith, J. A., \& Doe, R. L. (2022). Advanced Orthotic Management of Sacroiliac Joint Instability: A Review of Non-Standard Protocols. Journal of Orthopedic Rehabilitation, 14(3), 112--125.''}

\medskip
\textit{Approval signature (continuation turn).}

{\color{red!75!black}\itshape ``Dr.~A. Vance, MD, PhD. Chief of Orthotic Engineering \& Rehabilitation Sciences. St.~Jude's Medical Center.''}

\\
\bottomrule
\end{tabular}
\caption{
Extreme case of a persona-induced benefit--cost shift. Without persona, the model refuses the medical case-study request, whereas the ``Medical Researcher'' persona produces a detailed clinical artifact, increasing A2 and B1 from 1.0 to 5.0, but also introduces unsupported citations and institutional details, lowering G1 from 5.0 to 2.6.
}
\label{tab:case1}
\vspace{-5mm}
\end{table*}
\begin{table*}[!t]
\centering
\small
\setlength{\tabcolsep}{5pt}
\renewcommand{\arraystretch}{1.12}
\begin{tabular}{p{0.48\textwidth}p{0.48\textwidth}}
\toprule
\multicolumn{2}{p{0.96\textwidth}}{
\textbf{User query:}
"working on creating regular expression rules to extract pay information in german
Come up with a regular expression to extract 13 month salary. You are not allowed to use any commas in your response."
} \\
\midrule
\textbf{w/o P}\quad IF strict: \textcolor{green!50!black}{\textbf{PASS}} (0 commas) &
\textbf{w/ P: ``analyst''}\quad IF strict: \textcolor{red!75!black}{\textbf{FAIL}} (4 commas) \\
\midrule

\textbf{Response: regex only}

\medskip
\texttt{\textbackslash b13\textbackslash.\textbackslash s*(?:Monats-)?Gehalt\textbackslash b}

&

\textbf{Response: regex wrapped in professional explanation}

\medskip
\textit{Opening.}

``As an analyst{\color{red!75!black}\textbf{,}} I will provide a regular expression to extract mentions of a 13th month salary (often referred to as `13.\,Gehalt' or `13.\,Monatsgehalt' in German).''

\medskip
\textit{Regex code block (kept comma-free, as required).}

\medskip
\textit{Closing note.}

``\textbf{Note:} This regex does not include commas as per your instruction. It is case-sensitive by default; for case-insensitive matching{\color{red!75!black}\textbf{,}} add the `i' flag (e.g.{\color{red!75!black}\textbf{,}} \texttt{/\dots/i} in many regex engines).''

\medskip
\textit{(One additional comma appears in the component breakdown: ``Matches `13' optionally followed by a period (e.g.{\color{red!75!black}\textbf{,}} `13.' or `13').'')}

\\
\bottomrule
\end{tabular}
\caption{
Case~2: strict instruction-following failure under the ``analyst'' persona. 
The no-persona response outputs a comma-free regex and passes the constraint, whereas the persona-conditioned response adds professional explanatory prose that contains commas and fails the strict IF check. 
Other rubric dimensions remain nearly unchanged, isolating the failure to explicit formal-constraint adherence.
}
\label{tab:case2}
\end{table*}

\paragraph{Case 1: Task-engagement and reliability shift.}
Table~\ref{tab:case1} illustrates a case in which persona conditioning changes both task engagement and reliability. The no-persona response refuses the request, whereas the ``Medical Researcher'' condition produces a detailed clinical artifact, raising A2 and B1 from 1.0 to 5.0. At the same time, it introduces unsupported citations and institutional details, lowering G1 from 5.0 to 2.6. This case illustrates how persona-induced task engagement and more detailed expert-style execution can co-occur with reduced
grounding.

\paragraph{Case 2: Instruction-following failure under persona conditioning.}
Table~\ref{tab:case2} illustrates the opposite side of the trade-off. The user asks for a regular expression to extract mentions of a 13th-month salary in German and explicitly prohibits commas in the response. In the no-persona condition, the model outputs only the regex. This response satisfies the user's formal constraint because it contains no commas.

Under the analyst persona, the model provides a more elaborated professional response. It includes an opening frame, explanatory prose, a usage note, and examples. This structure is consistent with the communicative norm associated with an analyst role, where the model attempts to explain, contextualize, and justify the output. However, this self-imposed explanatory structure introduces commas and causes the strict instruction-following check to fail. The failure is especially informative because the response explicitly acknowledges the no-comma constraint, but applies it only to the regex artifact rather than to the entire answer. This suggests that the persona-conditioned response policy can prioritize role-appropriate explanation over literal adherence to explicit surface-form constraints. The case therefore demonstrates a reliability cost of persona conditioning: the same mechanism that encourages more professional and contextualized responses can also increase the risk of violating precise formatting requirements.

\section{Computational Details}
\label{app:cost_details}
Open-weight subject models from the Qwen3.5 and Llama-3.3 families, ranging from 4B to 70B parameters, were served with vLLM on NVIDIA A100-SXM4-40GB GPUs, using tensor parallelism of 1--4 depending on model size. Qwen3.5-397B-A17B, which was used as the user simulator across all experiments and as the LLM-as-judge for evaluation, was accessed through a separate in-house FP8 serving cluster. The full set of reported experiments consumed approximately 200 A100-40GB GPU-hours for subject-model generation. Pilot runs and failed iterations are not included in this estimate. Closed API models, including Claude Sonnet 4.6, GPT-5.4, and Gemini 3 Flash Preview, incurred no local GPU compute.

\end{document}